\documentclass[twocolumn]{main}

\paperstyle{fancy}
\papercolor{green}

\usepackage{cuted}
\usepackage{stfloats}
\usepackage{currfile}
\newcolumntype{L}{>{\raggedright\arraybackslash}X}
\newcolumntype{R}{>{\raggedleft\arraybackslash}X}
\definecolor{PaperTableHeader}{HTML}{F6E3DF}
\definecolor{PaperTableResult}{HTML}{FDF8E3}
\definecolor{PaperTableStripe}{HTML}{F2F2F2}
\definecolor{PaperGain}{HTML}{F26B3A}
\definecolor{PaperLoss}{HTML}{00B8D9}
\definecolor{PaperFlat}{HTML}{7A7A7A}
\newcommand{\TableBest}[1]{\textbf{#1}}
\newcommand{\TableSecond}[1]{\underline{#1}}
\newcommand{\TableUp}[1]{\textsuperscript{\textcolor{PaperGain}{$\uparrow$#1}}}
\newcommand{\TableDown}[1]{\textsuperscript{\textcolor{PaperLoss}{$\downarrow$#1}}}

\makeatletter
\newif\ifPaperBandPending
\newif\ifPaperFirstCell
\newif\ifPaperInTabular
\newif\ifPaperSuppMode
\PaperSuppModefalse
\newcommand{\PaperTableBands}{}
\newcommand{\PaperWideTableBands}{}
\newcommand{\EnablePaperTableBands}{%
  \ifPaperSuppMode
    \aboverulesep=0pt\belowrulesep=0pt
    \def\PaperTableBands{%
      \ifPaperInTabular\else
        \PaperInTabulartrue
        \global\PaperFirstCelltrue
        \rowcolors{1}{PaperTableHeader}{PaperTableHeader}%
        \let\PaperToprule\toprule
        \def\toprule{\noalign{\global\rownum\@ne\global\PaperFirstCelltrue}\showrowcolors\PaperToprule}%
        \let\PaperMidrule\midrule
        \def\midrule{\hiderowcolors\noalign{\global\PaperFirstCelltrue}\PaperMidrule}%
        \let\PaperArrayCr\@arraycr
        \def\@arraycr{\PaperArrayCr\noalign{\global\PaperFirstCelltrue}}%
        \let\PaperCellPaint\CT@@do@color
        \def\CT@@do@color{%
          \ifPaperFirstCell
            \global\PaperFirstCellfalse
          \else
            \advance\@tempdimb 0.25pt\relax
          \fi
          \PaperCellPaint}%
      \fi}%
    \def\PaperWideTableBands{\PaperTableBands}%
  \else
    \aboverulesep=0pt\belowrulesep=0pt
    \def\PaperTableBands{%
      \rowcolors{1}{PaperTableHeader}{PaperTableHeader}%
      \let\PaperMidrule\midrule
      \def\midrule{\hiderowcolors\PaperMidrule}}%
    \def\PaperWideTableBands{%
      \PaperTableBands
      \rowcolors[\noalign{\global\PaperBandPendingtrue}]{1}{PaperTableHeader}{PaperTableHeader}%
      \let\PaperCellPaint\CT@@do@color
      \def\CT@@do@color{%
        \ifPaperBandPending
          \global\PaperBandPendingfalse
          \rlap{\vrule width\linewidth height\ht\@arstrutbox depth\dp\@arstrutbox}%
        \fi
        \@tempdimb\z@ \@tempdimc\z@
        \PaperCellPaint}}%
  \fi}
\AtBeginDocument{%
  \patchcmd{\@@@cmidrule}
    {\ifnum\@lastruleclass=\z@\vskip \aboverulesep\fi}
    {\ifnum\@lastruleclass=\z@\vskip \aboverulesep
       \ifPaperSuppMode
         \if@rowcolors
           \vskip-0.15pt
           {\color{PaperTableHeader}\hrule \@height\dimexpr\@thisrulewidth+0.3pt\relax}%
           \vskip-\dimexpr\@thisrulewidth+0.15pt\relax
         \fi
       \fi
     \fi}
    {}{}%
}
\makeatother
\AtBeginEnvironment{table}{\EnablePaperTableBands}
\AtBeginEnvironment{table*}{\EnablePaperTableBands}
\AtBeginEnvironment{tabular}{\PaperTableBands}
\AtBeginEnvironment{tabular*}{\PaperWideTableBands}
\AtBeginEnvironment{tabularx}{\PaperTableBands}

\newcommand{\numnVR}{1,178}

\newcommand{\numscSB}{64.4}

\newcommand{\numscSBnone}{63.4}

\newcommand{\numpairOurs}{33.3}

\newcommand{\numpairTCe}{19.7}

\newcommand{\numpairSlide}{32.1}

\newcommand{\cdBrho}{0.23}
\newcommand{\cdBratio}{1.12}

\newcommand{\cdBtum}{0.24}

\newcommand{\cdErho}{0.53}
\newcommand{\cdEratio}{4.44}

\newcommand{\cdEtum}{0.52}
\newcommand{\cdBmean}{0.293}
\newcommand{\cdEmean}{1.234}

\newcommand{\cdBngt}{7}
\newcommand{\cdEngt}{13}
\newcommand{\cdNattr}{13}

\newcommand{\cdBrhoAttn}{0.07}
\newcommand{\cdErhoAttn}{0.31}

\newcommand{\regionalNTransplant}{1,124}

\newcommand{\GtwoAuxTr}{10.54}
\newcommand{\GtwoPairedPair}{34.33}

\newcommand{\GtwoFullTr}{22.48}
\newcommand{\GtwoGainDelta}{8.44}
\newcommand{\GtwoGainLo}{4.90}
\newcommand{\GtwoGainHi}{12.44}

\newcommand{\caseMainOriginal}{83.9}
\newcommand{\caseMainSingle}{3.6}

\newcommand{\caseMainBaselineOriginal}{55.0}
\newcommand{\caseMainBaselineSingle}{54.9}
\newcommand{\caseMainToken}{4}
\newcommand{\caseFailureOriginal}{90.5}
\newcommand{\caseFailureSingle}{85.6}
\newcommand{\caseFailureAll}{87.9}

\newcommand{\auxWSIBeforeMean}{48.0}

\newcommand{\auxWSIAfterMean}{60.0}

\newcommand{\artifactN}{1,384}
\newcommand{\artifactEvaTumour}{31.4}

\newcommand{\artifactEvaContext}{6.6}

\newcommand{\vtN}{48}

\newcommand{\vet}{VET\xspace}
\newcommand{\ours}{PathSleuth\xspace}

\newcommand{\maketitlesupplementary}{%
  \clearpage
  \twocolumn[%
    \noindent{\Large\sffamily\bfseries\color{accentcolor}Appendix\par}%
    \vspace{0.25em}%
    {\color{rulecolor!70}\rule{\textwidth}{0.65pt}}\par%
    \vspace{0.6em}%
  ]%
  \addcontentsline{toc}{section}{Appendix}%
}

\title{Guessing or Looking? Visual Evidence Tracing and Anchoring for Whole-Slide Pathology VLMs}

\author[1]{Zhuo Chen}
\author[1,\ast]{Wenshuo Chen}
\author[1]{Shaofeng Liang}
\author[2]{Lei Wang}
\author[3]{Le Yang}
\author[1]{Zirui Wang}
\author[1]{Kaishen Yuan}
\author[1]{Songning Lai}
\author[4]{Jin-Hao Sheng}
\author[1]{Lujundong Li}
\author[1,\dagger]{Yutao Yue}

\affiliation[1]{The Hong Kong University of Science and Technology (Guangzhou)}
\affiliation[2]{Griffith University}
\affiliation[3]{Sun Yat-sen University}
\affiliation[4]{Nanjing University}

\contribution[\ast]{Project leader}
\contribution[\dagger]{Corresponding author}
\correspondence{\email{yutaoyue@hkust-gz.edu.cn}}

\abstract{
Whole-slide pathology questions often hinge on local tissue patterns, such as those distinguishing ductal from lobular breast carcinoma.
Models compress gigapixel slides into a few visual tokens. Answer accuracy alone gives little indication of which tissue regions influenced a prediction.
PathSleuth uses Visual Evidence Tracing for visual attribution across tissue patches, eight compressed slots and the language model.
We trace each slot to its patches through routing weights and test their influence by gradient-guided deletion.
Linear probes measure how well each slot encodes the queried attribute. Replacing one slot at a time with its counterpart from another patient measures its effect on the correct answer.
We compare attention across heads and layers at question and pre-answer positions, then block visual connections at individual layers.
In TC-SSA-8, attribute information correlates weakly with slot influence.
PathSleuth combines report-attribute supervision with evidence-anchored instruction tuning on slide substitutions and regional feature exchanges that preserve or change the target.
After training, slots that rank highly across attributes have greater influence on answers. Restoring the three highest-ranked slots after a full swap recovers 81\% of the mean loss in correct-answer log probability.
PathSleuth-8 leads the evaluated models on both SlideBench-Evidence measures, while its SlideChat-initialised variant leads on SlideBench-VR.
}

\hypersetup{pdfauthor={Zhuo Chen, Wenshuo Chen, Shaofeng Liang, Lei Wang, Le Yang, Zirui Wang, Kaishen Yuan, Songning Lai, Jin-Hao Sheng, Lujundong Li, Yutao Yue}}

\begin{document}
\maketitle
\begin{figure}[t]
\centering
\includegraphics[width=\linewidth]{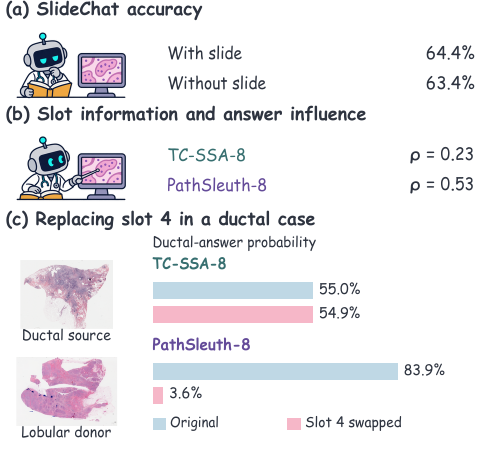}
\caption{\textbf{What reaches the answer?} (a) SlideBench accuracy. (b) Mean rank correlation of probe AUC and slot effect across 13 attributes (780 questions). (c) Ductal-answer probability before and after replacing slot 4 with the lobular donor's slot.}
\label{fig:motivation}
\end{figure}

\section{Introduction}
\label{sec:intro}

Consider two patients with breast cancer, one ductal and one lobular. Both slides prompt the same subtype question. The organ narrows the plausible answers for both patients, but the distinction requires evidence from each slide. In our SlideBench evaluation, SlideChat~\citep{Chen2025SlideChat} answers \numscSB\% of questions correctly with the patient's slide and \numscSBnone\% without it (\cref{fig:motivation}a). Even a reader that changes its answer when the slides are exchanged can assign the wrong subtype to both patients. How can we tell which tissue evidence contributed to either answer?

Compression can obscure patient evidence. Our eight-slot TC-SSA baseline~\citep{Chen2026TCSSA} pools thousands of tissue patches before seeing the question, so the same slots must support a question about subtype and one about lymphovascular invasion. If pooling obscures a small focus of invasion, the reader has little evidence for that finding. A different failure is possible when the relevant information survives compression. A slot may distinguish ductal from lobular carcinoma across patients, even if changing that slot barely affects the reader's subtype prediction. These two failures call for different changes to the model, depending on whether tissue aggregation has obscured the finding or the reader has overlooked it.

The ductal case in \cref{fig:motivation}c illustrates how to isolate one slot's influence. We keep the question, options and seven slots fixed, then replace slot~\caseMainToken{} with its counterpart from a lobular case. The probability of the ductal answer drops from \caseMainOriginal\% to \caseMainSingle\% in \ours, while the same substitution moves it from \caseMainBaselineOriginal\% to \caseMainBaselineSingle\% in TC-SSA-8. We chose this position because it produced the largest drop for the ductal case in \ours. An influential slot in this patient might carry several features that differ between the two slides. To ask whether it consistently distinguishes the reported subtype, we need a second measurement across patients. Visual Evidence Tracing (\vet) makes that comparison for each report attribute. A linear probe ranks the eight slots by how well each distinguishes the attribute's labels in held-out patients. We independently rank the same slots by the magnitude of the change in correct-answer log probability when each is replaced. We test the probe ranking in reverse by starting with eight donor slots and restoring original slots in probe order or at random.

The breast cases also pose a spatial question. An attention map might highlight tumour in both slides even if the reader uses the highlighted tissue only to recognise carcinoma. Routing weights tell us which patches enter each slot. Across the slide, we rank patches by how sensitive the predicted answer's log probability is to their inclusion. We then remove the highest-ranked patches, recompute the slots, and compare the change in that prediction with random deletion. Inside the reader, we measure attention to the slots from the question and pre-answer positions. Blocking those connections one layer at a time tests where the answer depends on the incoming visual information. In TC-SSA-8, the best-slot probe AUC averages 0.75 across 13 report attributes. The slot ranking by probe AUC correlates only weakly with the ranking by influence on the answer (mean Spearman $\rho=\cdBrho$, \cref{fig:motivation}b).

A reader trained on original questions alone gets credit for the correct subtype regardless of which slot supplied the evidence. For the breast cases, the missing supervision is what should happen when the reader sees another patient's tissue under exactly the same question. The reports supply this distinction for training even when they do not identify the diagnostic patch. This observation motivates the two training signals in \ours. Report labels supervise what the visual slots retain, while evidence-anchored instruction tuning (EvA) uses paired slides to teach the reader when the same question should have a different answer. A second ductal case should retain the ductal target even though its tissue looks different, whereas a lobular case should change the target under the same question and options. For a local change, we exchange tumour-associated features between slides, retain the recipient's surrounding tissue and train against the donor label. At inference the reader receives only the patient's slide and question.

At evaluation, we return to the two original patients. A reader that always chooses the more common breast subtype can earn credit on individual slides. Requiring both diagnoses to be correct tests whether the reader can distinguish patients who share the same cancer type. In a separate test we keep the slide fixed and rotate the options. The diagnosis stays the same, but its letter changes with each rotation.

\begin{itemize}
\setlength{\itemsep}{1pt}\setlength{\parskip}{0pt}
\item \textbf{Patient-level evaluation.} SlideBench-VR screens for text-only solvability and tests four option orders. SlideBench-Evidence measures how often the reader answers both patients' original questions correctly when their reports give different labels.
\item \textbf{Visual attribution.} After training, the slot ranking by report-attribute probe AUC agrees more closely with the ranking by answer influence (mean Spearman $\rho=\cdErho$ for \ours-8).
\item \textbf{Evidence-anchored training.} With the same data and one instruction epoch, training the visual modules with attribute heads and whole-slide targets raises paired correctness from \numpairTCe\% to \numpairSlide\%. The full two-epoch recipe reaches \numpairOurs\%.
\end{itemize}

\section{Related Work}
\label{sec:related}

\paragraph{Whole-slide representations and supervision.}
Pathology encoders represent tissue at patch and slide scales~\citep{Lu2024VisualLanguage,Chen2024Towards,Xu2024WholeSlide,Ding2024TITAN} for whole-slide question answering~\citep{Chen2025SlideChat,Liang2025WSILLaVA}. BLIP-2 uses learned queries to bridge frozen encoders and language models~\citep{Li2023BLIP2}. TC-SSA pools patches into fixed slots~\citep{Chen2026TCSSA}, while LoC-Path and PathSelect choose features for each question~\citep{Hu2025LoCPath,Chen2026PathSelect}. Counterfactual approaches mask training inputs~\citep{Chen2020CSS} or subtract the question’s direct effect at inference~\citep{Niu2021CounterfactualVQA}. Pair-DPO learns preferences over counterfactual slide pairs~\citep{Zhang2026CleanSlide}. EvA supervises eight slots under slide and region exchanges.

\paragraph{Visual dependence and attribution.}
Paired images~\citep{Goyal2017VQAv2,Li2024NaturalBench,Tong2024MMVP} and shifts in answer priors~\citep{Agrawal2018VQACP} test visual dependence. PathoArgus replaces or removes slides under fixed questions~\citep{Liu2026PathoArgus}. MIRAGE tests medical QA without images~\citep{Asadi2026Mirage}. PathBind tests visual--semantic binding~\citep{Zhang2026PathBind} and EviPathBench tests evidence acquisition~\citep{Liao2026EviPathBench}. Readers can overlook decodable features~\citep{Fu2025HiddenPlainSight,Wang2026DermUnderutilize}. SCOPE relates slide predictions to tissue concepts~\citep{Lazard2026SCOPE}. We use probes~\citep{Alain2017Probes} and activation patching~\citep{Meng2022ROME,Zhang2024Patching} to test whether attribute-encoding slots influence the corresponding answers.

\begin{figure*}[!t]
\centering
\includegraphics[height=0.65757\textwidth]{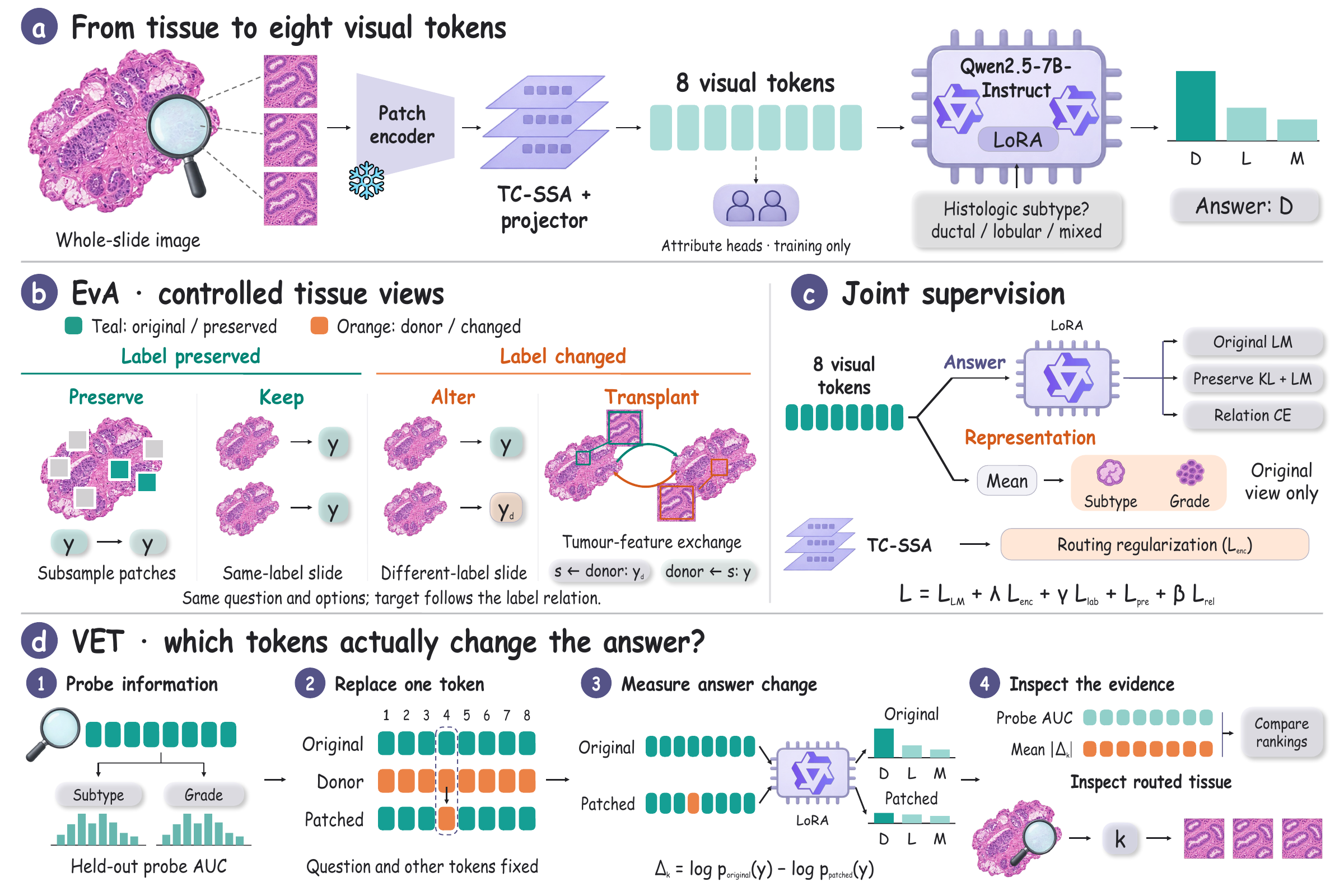}
\caption{\textbf{PathSleuth training and evidence tracing.} (a) Eight visual tokens connect the frozen patch encoder to the reader. (b) Preserve and Keep retain the label, while Alter and Transplant change it. (c) Answer and attribute losses supervise the reader and tokens. (d) VET probes each token for an attribute and replaces it with a donor token to measure its effect on the answer.}
\label{fig:method}
\end{figure*}

\section{From Answer Accuracy to Evidence Use}
\label{sec:diagnosis}
\subsection{Controlled answer comparisons}
\label{sec:rootcause}
We vary option order and slide identity separately. The \href{https://huggingface.co/datasets/General-Medical-AI/SlideChat/blob/aa1aa40b78564b094e3c3305918ec7f7fc1ba5af/SlideBench-VQA-TCGA.csv}{expanded SlideBench-VQA-TCGA release}~\citep{Chen2025SlideChat} contains 3,176 questions across 31 cancer types. We score each question in four cyclic option orders and report their mean accuracy and the fraction correct in \emph{all} orders (CircularEval~\citep{Liu2024MMBench}). SlideBench-VR retains \numnVR{} questions for which none of four fixed text-only judges succeeds in every order, after excluding answers contradicted by the report audit.

SlideBench-Evidence contains 3,994 questions about report attributes on 1,425 slides. Each eligible slide is paired with a different patient of the same cancer type whose report gives a different value for the queried attribute. \emph{Both correct} requires the reader to answer each slide's original question correctly. \emph{Keep} measures accuracy after a same-label replacement. The pair tests whether the reader distinguishes the patients, while Keep asks whether it recognises a finding they share. Filter sensitivity, question construction and text-only controls appear in \cref{sec:app_vr,sec:app_data,tab:certify}.

\subsection{Visual Evidence Tracing}
\label{sec:vet_main}
\Cref{fig:method}d shows how \vet connects token information to answer influence. Attribute
probes read individual compressed slots and measure report-label decodability by held-out
AUC. Token substitutions then measure the reader's response with the question fixed.

For a question $q$ about attribute $a$ on slide $s$, we select another slide $\bar s$ with a
different report label for $a$, matching the cancer type whenever possible. Let $V_s=(v_{s,1},\ldots,v_{s,K})$ denote the original $K$ visual tokens, with $k$
indexing a token and $v_{s,k}$ its vector. Write $V_s^{\,k\leftarrow\bar s}$ for the sequence
after replacing token $k$ with its counterpart from $\bar s$. Let $\pi_\theta(y\mid q,V)$ be the reader probability for answer $y$ given question $q$
and tokens $V$, with $\theta$ denoting model parameters. Following activation patching~\citep{Meng2022ROME,Zhang2024Patching}, we measure the
change in log-probability assigned to the original correct answer $y$:
\begin{equation}
\Delta_k(q,s) = \log \pi_\theta\big(y \mid q, V_s\big) - \log \pi_\theta\big(y \mid q, V_s^{\,k\leftarrow \bar s}\big),
\label{eq:patch}
\end{equation}
For each attribute, we rank the eight slots by probe AUC and by mean $|\Delta_k|$ across its questions, then compute Spearman's $\rho$ between these rankings. We average the correlations over the 13 attributes. A high value means that the slots that best distinguish an attribute's labels also have the largest effects on answers about it.

Ranking slots by the amount of tumour tissue they collect gives a different result. On \vtN{} questions, blocking reads to the two slots with the most tumour-patch mass lowers correct-answer log probability by 0.14 nats for TC-SSA-8 and 0.03 for \ours. Tumour content alone does not identify which aspect of a tumour the question asks about. We use the attribute-specific probe ranking for \vet and report the tissue and attention controls separately (\cref{sec:app_vet,tab:vision_text}).

\section{Evidence-Anchored Instruction Tuning}
\label{sec:method}
\ours constructs related visual inputs with known targets, encodes them through a shared
visual interface, and jointly supervises their token representations and answers (\cref{fig:method}).

\textbf{Preliminaries.}
Let $s$ index a slide and $X_s\in\mathbb{R}^{N_s\times D}$ contain its $N_s$ patch features
of dimension $D$. A slide encoder $E$ and projector $g$ produce
$V_s=g(E(X_s))=(v_{s,1},\ldots,v_{s,K})\in\mathbb{R}^{K\times H}$, where $K$ is the
visual-token budget and $H$ is the reader's embedding dimension. Unsubscripted $X,V$
denote any input view and its tokens. Given question $q$, the
reader assigns probability $\pi_\theta(y\mid q,V)$ to answer $y$; $\theta$ denotes the
trainable parameters. For a multiple-choice question, $p_V=\pi_\theta(\cdot\mid q,V)$
is normalised over its valid options, and $\ell(y,V)=-\log p_V(y)$ is option-level
cross-entropy. We use $\mathcal{L}_{\mathrm{LM}}(y\mid q,V)$ for autoregressive
cross-entropy averaged over supervised answer tokens. \Cref{tab:notation} summarises the notation.

All views share $E$, $g$ and the reader; the patch extractor is fixed.
In \cref{alg:eva}, $\operatorname{ViewTarget}$ constructs relation inputs $X_i^{(r)}$
and targets $y_i^{(r)}$ under fixed options (\cref{sec:rea}). The accumulation count $m$
sets the scale of each micro-batch gradient; $\theta.\mathrm{grad}$ stores their sum.

\label{sec:evis}

\subsection{Visual views and answer targets}
\label{sec:rea}
Let $a$ index a report attribute, $y_{s,a}$ its label for slide $s$, and $\mathcal{Y}_a$
the candidate labels. A shared question stem $t_a$ and an option permutation $\sigma$ define
\begin{equation}
q_{s,a}=\big(t_a,\sigma(\mathcal{Y}_a)\big),\qquad \text{target }y_{s,a}.
\label{eq:question}
\end{equation}
Each label denotes its corresponding answer letter under $\sigma$. Within each pair we fix the stem and option order in \cref{eq:question}, so changed targets cannot be explained by changed text.
Report extraction supplies the labels; the following relations determine the supervision
for each visual input (\cref{sec:app_data}).

\emph{Preserve} retains each patch with probability $\rho\in(0,1]$, keeping at least one
valid patch, to form $X'_s$ and $V'_s=g(E(X'_s))$. The original answer remains the target.
We also match the full-slide option distribution:
\begin{equation}
\begin{split}
\mathcal{L}_{\text{pre}} &= \mathrm{KL}\!\left(\mathrm{sg}[p_{V_s}]\,\|\,p_{V'_s}\right)\\[4pt]
&\quad + \mu\,\mathcal{L}_{\text{LM}}(y\mid q,V'_s).
\end{split}
\label{eq:pre}
\end{equation}
Here $\mathrm{KL}$ is Kullback-Leibler divergence, $\mathrm{sg}$ stops gradients through
its argument, and $\mu\geq0$ weights subsampled-answer supervision. Only the KL teacher is
detached: original-answer supervision still trains the full-slide branch, and both views
update shared parameters. Preserve applies to closed-form questions, whereas the remaining
relations require report-derived targets.

\emph{Alter} selects another patient's slide $\bar s$ with $y_{\bar s,a}\ne y_{s,a}$;
\emph{Keep} selects $\tilde s$ with $y_{\tilde s,a}=y_{s,a}$. Both prefer the same cancer
type and require targets among the fixed options. Their losses are
\begin{equation}
\begin{split}
\mathcal{L}_{\text{alt}}&=\ell(y_{\bar s,a},V_{\bar s}),\\
\mathcal{L}_{\text{keep}}&=\ell(y_{s,a},V_{\tilde s}).
\end{split}
\label{eq:alt}
\end{equation}
The queried labels determine whether replacement should change the answer: Alter changes
the target, whereas Keep preserves it.

On an Alter pair the question is identical and the two labels differ, so a reader using only text incurs a summed unweighted log loss of at least $2\log 2$, attained by assigning each label probability one half.

\emph{Transplant} is Alter applied to a region rather than the whole slide. Let $M_s$ be the
set of patch indices selected by a tumour locator, and $X_s[\neg M_s]$ the unselected
features. Draw $\widehat X_{\bar s\to s}$ by sampling $|M_s|$ features from $X_{\bar s}[M_{\bar s}]$.
Set $X_{s\leftarrow\bar s}[M_s]=\widehat X_{\bar s\to s}$ and
$X_{s\leftarrow\bar s}[\neg M_s]=X_s[\neg M_s]$. Its representation is
$V_{s\leftarrow\bar s}=g(E(X_{s\leftarrow\bar s}))$. Exchanging both directions gives
\begin{equation}
\begin{split}
\mathcal{L}_{\text{tr}} &= \ell\!\left(y_{\bar{s},a},V_{s\leftarrow\bar{s}}\right)\\[4pt]
&\quad + \ell\!\left(y_{s,a},V_{\bar{s}\leftarrow s}\right).
\end{split}
\label{eq:tr}
\end{equation}
The recipient keeps its context and patch count. The donor supplies the synthetic target for tumour-associated attributes when both slides have eligible regions. Training and testing share a locator. Its tumour and necrosis fractions correlate positively with pathologists' estimates on matched frozen sections in 78\% and 94\% of cancer types, respectively (\cref{sec:app_locator}).

\begin{table*}[t]\centering\small
\caption{\textbf{Answer consistency and paired correctness (\%).} VR accuracy averages four cyclic orders. All orders requires four correct answers. Both correct requires correct answers to both original questions of a contrastive pair. TC-SSA-8 and \ours-8 each use eight visual tokens.}
\label{tab:main}
\begin{tabular*}{\textwidth}{@{\extracolsep{\fill}}l cc cc r@{}}\toprule
& \multicolumn{2}{c}{SlideBench-VR} & \multicolumn{2}{c}{SlideBench-Evidence} & Visual \\
\cmidrule(lr){2-3}\cmidrule(lr){4-5}
Method & Accuracy & All orders & Accuracy & Both correct & tokens \\\midrule
\rowcolor{PaperTableStripe}
Text-only reader & 36.1\TableDown{5.5} & 10.6\TableDown{8.1} & 51.9\TableDown{8.0} & 5.6\TableDown{14.1} & 0\\
\midrule
\rowcolor{white}
\multicolumn{6}{l}{\emph{Released checkpoints (original training data)}} \\
\rowcolor{white}
SlideChat$^\dagger$ & 39.9\TableDown{1.7} & 8.5\TableDown{10.2} & 40.6\TableDown{19.3} & 13.7\TableDown{6.0} & 15.9k\\
\rowcolor{PaperTableStripe}
SLICEChat$^{\dagger\ddagger}$ & 44.9\TableUp{3.3} & 17.8\TableDown{0.9} & 54.0\TableDown{5.9} & 20.1\TableUp{0.4} & 705\\
\rowcolor{white}
HistoSelect$^{\dagger\|}$ & 38.4\TableDown{3.2} & 11.9\TableDown{6.8} & 46.6\TableDown{13.3} & 5.2\TableDown{14.5} & 2.1k\\
\midrule
\rowcolor{white}
\multicolumn{6}{l}{\emph{Common $\mathcal{D}_{\mathrm{ev+}}$ data and base Qwen reader}} \\
\rowcolor{PaperTableStripe}
Mean pooling & 38.4\TableDown{3.2} & 11.7\TableDown{7.0} & 54.8\TableDown{5.1} & 12.9\TableDown{6.8} & 1\\
\rowcolor{white}
TC-SSA-8 & 41.6 & 18.7 & 59.9 & 19.7 & 8\\
\rowcolor{PaperTableStripe}
TCP-LLaVA$^{\ast\S}$ & 35.4\TableDown{6.2} & 5.2\TableDown{13.5} & 51.0\TableDown{8.9} & 5.6\TableDown{14.1} & 100\\
\rowcolor{white}
LoC-Path$^\ast$ & 40.2\TableDown{1.4} & 13.2\TableDown{5.5} & 57.0\TableDown{2.9} & 20.8\TableUp{1.1} & 96\\
\rowcolor{PaperTableStripe}
SparseLearn / PathSelect$^\ast$ & 41.9\TableUp{0.3} & 18.4\TableDown{0.3} & 59.2\TableDown{0.7} & 19.2\TableDown{0.5} & 32\\
\midrule
\rowcolor{white}
LoC-Path + \ours & 49.2\TableUp{7.6} & 28.1\TableUp{9.4} & 63.4\TableUp{3.5} & \TableSecond{32.7}\TableUp{13.0} & 96\\
\rowcolor{PaperTableResult}
\ours-8 & \TableSecond{49.6}\TableUp{8.0} & \TableSecond{31.0}\TableUp{12.3} & \TableBest{64.7}\TableUp{4.8} & \TableBest{33.3}\TableUp{13.6} & 8\\
\midrule
\rowcolor{white}
\multicolumn{6}{l}{\emph{Additional SlideChat reader initialization, permutation and 25k replay}} \\
\rowcolor{white}
\ours-SC (E3)$^\diamond$ & \TableBest{54.0}\TableUp{12.4} & \TableBest{42.0}\TableUp{23.3} & \TableSecond{63.6}\TableUp{3.7} & 30.1\TableUp{10.4} & 8\\
\bottomrule\end{tabular*}
\par\smallskip\begin{minipage}{\textwidth}\footnotesize $^\dagger$Public weights; $^\ast$visual-module re-implementations, with differences listed in \cref{sec:app_baselines}; $^\ddagger$CONCH-v1.5, 96--97\% coverage; $^\S$10,000-patch cap; $^\|$question-dependent budget. $^\diamond$Combined recipe, not an initialization ablation. \ours-8 supplies the evidence tracing. Arrows show point differences from TC-SSA-8. Bold and underline mark first and second, including ties.\end{minipage}\end{table*}

\begin{algorithm}[t]
\caption{Evidence-anchored instruction tuning}
\label{alg:eva}
\small
\begin{algorithmic}[1]
\algtext*{EndFor}
\algtext*{EndIf}
\Require Micro-batch $B$, parameters $\theta$, accumulation count $m$
\State $V_s\gets g(E(X_s))$ for all $s\in B$
\State $\mathcal{L}_{\mathrm{rel}}\gets0$
\For{$r\in\{\mathrm{alt},\mathrm{keep},\mathrm{tr}{\to},\mathrm{tr}{\leftarrow}\}$}
  \State $I_r\gets\{i\in B:\text{relation }r\text{ is eligible}\}$
  \If{$|I_r|>0$}
    \State $(X_i^{(r)},y_i^{(r)})_{i\in I_r}\gets\operatorname{ViewTarget}(I_r,r)$
    \State $V_i^{(r)}\gets g(E(X_i^{(r)}))$ for all $i\in I_r$
    \State $\mathcal{L}_{\mathrm{rel}}\gets\mathcal{L}_{\mathrm{rel}}+\frac{1}{|I_r|}\sum_{i\in I_r}\ell(y_i^{(r)},V_i^{(r)})$
  \EndIf
\EndFor
\State $\mathcal{L}\gets\text{Eq.~\eqref{eq:total}}$
\State $\theta.\mathrm{grad}\gets\theta.\mathrm{grad}+\nabla_\theta\mathcal{L}/m$
\If{accumulation boundary reached}
  \State $\theta\gets\operatorname{OptimizerStep}(\theta)$; $\theta.\mathrm{grad}\gets0$
\EndIf
\end{algorithmic}
\end{algorithm}

\subsection{Joint representation and answer supervision}
\label{sec:tokens}
The answer objectives act through the reader, while attribute supervision directly constrains
what the visual representation retains. Specifically, an auxiliary head $h_a$ returns a distribution over $\mathcal{Y}_a$ from the original-view mean
$\bar v_s=K^{-1}\sum_{k=1}^{K}v_{s,k}$. For a micro-batch $B$, let $B_a$ index its original
rows labelled for $a$, and $A_B=\{a:|B_a|>0\}$. With $\mathrm{CE}$ denoting categorical
cross-entropy, attribute supervision is
\begin{equation}
\mathcal{L}_{\text{lab}} = \frac{1}{|A_B|}\sum_{a\in A_B}\frac{1}{|B_a|}
\sum_{s\in B_a}\mathrm{CE}\!\left(h_a(\bar v_s),y_{s,a}\right),
\label{eq:lab}
\end{equation}
with value zero if $A_B$ is empty. The head predicts the report label independently of
question wording, while the reader selects the corresponding answer among the current options. Joint updates
to $E$, $g$ and the reader allow both objectives to shape the same visual tokens.

Relation availability varies across rows. For $r\in\{\mathrm{alt},\mathrm{keep},
\mathrm{tr}{\to},\mathrm{tr}{\leftarrow}\}$, let $I_r\subseteq B$ be its eligible rows,
and $(y_i^{(r)},V_i^{(r)})$ the target and visual representation of row $i$. Define
$\bar\ell_r=|I_r|^{-1}\sum_{i\in I_r}\ell(y_i^{(r)},V_i^{(r)})$ for nonempty $I_r$,
and zero otherwise. Then $\mathcal{L}_{\mathrm{rel}}=\sum_r\bar\ell_r$ sums separate
relation means, including both directions of regional Alter. Combining the view losses in \cref{eq:pre,eq:alt,eq:tr} gives
\begin{equation}
\begin{split}
\mathcal{L} &= \mathcal{L}_{\text{LM}} + \lambda\,\mathcal{L}_{\text{enc}} + \gamma\,\mathcal{L}_{\text{lab}}\\[4pt]
&\quad + \mathcal{L}_{\text{pre}} + \beta\,\mathcal{L}_{\text{rel}},
\end{split}
\label{eq:total}
\end{equation}
where $\mathcal{L}_{\mathrm{enc}}$ is the base encoder's regularisation loss and
$\lambda$, $\gamma$, $\beta\geq0$ weight encoder, attribute and relation supervision.
Here $\mathcal{L}_{\mathrm{LM}}$ uses original answers. In $\mathcal{L}_{\mathrm{pre}}$,
KL averages eligible rows and the LM term averages supervised answer tokens separately.
There is no additional division by the number of active relations.

\Cref{alg:eva} summarises the training step. The trainable parameters $\theta$ include
$E$, $g$, the reader's adapted parameters and $\{h_a\}$; the frozen patch extractor supplies
all views. At inference we discard the attribute heads and relation construction, using only
$X_s\mapsto V_s\mapsto\pi_\theta(\cdot\mid q,V_s)$. Architecture choices and numerical
coefficients are specified in \cref{sec:experiments,sec:app_impl}.

\begin{figure*}[t]
\centering
\includegraphics[width=\linewidth]{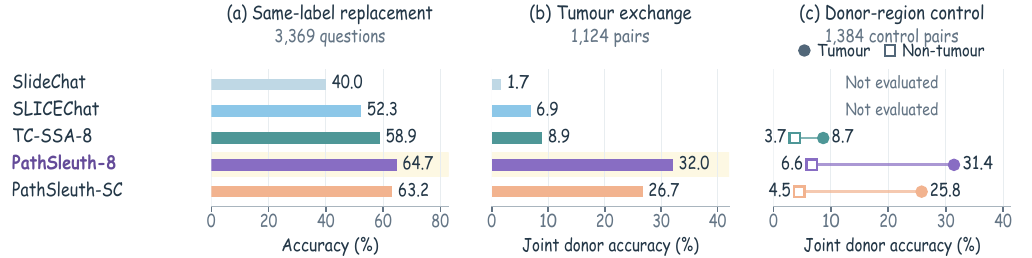}
\caption{\textbf{Answers after slide replacement and regional exchange.} (a) Accuracy after same-label replacement (3,369 questions). (b) Both views must follow the tumour donor (1,124 pairs). (c) Tumour versus non-tumour exchange on 1,384 control pairs. CONCH defines the exchanged regions.}
\label{fig:regional}
\label{tab:evidence}
\end{figure*}

\section{Experiments}
\label{sec:experiments}
\subsection{Counterfactual answer evaluation}
\label{sec:counterfactual_eval}
\ours uses eight TC-SSA slots~\citep{Chen2026TCSSA}, CONCH features and Qwen2.5-7B-Instruct~\citep{Qwen2024Qwen25} with rank-64 LoRA~\citep{Hu2022LoRA}. For \ours-8 we train for 15 epochs on captions and two epochs on 130,837 instructions, including 89.3k closed-form SlideChat examples and questions drawn from TCGA reports~\citep{Kefeli2024TCGAReports}. Evaluation patients are excluded from training questions and report-label pools. Other encoders retain their native token budgets (\cref{sec:app_impl,sec:app_baselines}).

Each pair in \cref{tab:main} retains both original questions. At the same eight-slot budget, \ours gains 8.0 points in VR accuracy and 12.3 in all-orders accuracy over the final TC-SSA-8 model. At one instruction epoch on the same data, training the visual modules with attribute heads and whole-slide supervision raises paired correctness from \numpairTCe\% to \numpairSlide\% (\cref{tab:ablation}). The two-epoch recipe reaches \numpairOurs\%. At 96 tokens, LoC-Path improves from 20.8\% to 32.7\% under the same recipe. SLICEChat changes more answers after a same-cancer swap and answers fewer pairs correctly (\cref{tab:main_detailed}).

\begin{figure*}[!b]
\centering
\includegraphics[width=\linewidth]{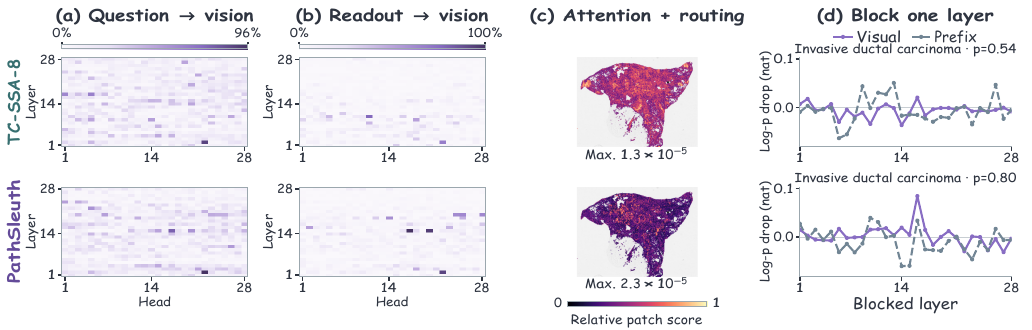}
\caption{\textbf{Visual reads in breast-subtype case 318.} (a,b) Question and pre-answer attention to the eight slots by head and layer. (c) Routing-weighted attention is diffuse for TC-SSA-8 and more focal for \ours on this slide. Each map uses its own maximum. (d) Drop in the original answer's log probability after blocking visual or length-matched prefix reads in one layer.}
\label{fig:attention_capture}
\label{fig:visual_evidence}
\end{figure*}

\Cref{fig:regional} tests whether answers follow exchanged tissue. We exchange CONCH-labelled tumour features~\citep{Lu2024VisualLanguage} in \regionalNTransplant{} eligible pairs and score both donor labels. On the separate \artifactN{} control pairs, \ours follows both donors in \artifactEvaTumour\% of tumour exchanges and \artifactEvaContext\% of non-tumour exchanges, a 4.8-fold difference with recipient context held fixed (\cref{tab:transplant_controls}).

\begin{table}[t]
\centering\footnotesize
\setlength{\tabcolsep}{2pt}
\caption{\textbf{Evidence tracing.} Slot swaps cover 780 questions, attention 150, and tumour correlation ten attributes. Top three refers to probe AUC.}
\label{tab:vet_focus}
\begin{tabular*}{\linewidth}{@{\extracolsep{\fill}}lcc@{}}
\toprule
 & TC-SSA-8 & \ours-8 \\
\midrule
\rowcolor{PaperTableStripe}
Full-slide swap, mean drop (nats) & \TableSecond{\cdBmean} & \TableBest{\cdEmean}\\
\rowcolor{white}
Tumour fraction--effect $\rho$ & \TableSecond{\cdBtum} & \TableBest{\cdEtum}\\
\rowcolor{PaperTableStripe}
Attention--effect $\rho$ & \TableSecond{\cdBrhoAttn} & \TableBest{\cdErhoAttn}\\
\rowcolor{white}
Information--effect $\rho$ & \TableSecond{\cdBrho} & \TableBest{\cdErho}\\
\rowcolor{PaperTableStripe}
Top-three / other effect & \TableSecond{\cdBratio} & \TableBest{\cdEratio}\\
\rowcolor{white}
Attributes with ratio $>1$ & \TableSecond{\cdBngt}/\cdNattr & \TableBest{\cdEngt}/\cdNattr\\
\bottomrule
\end{tabular*}
\end{table}

\subsection{Tracing evidence from patches to answers}
\label{sec:causal_main}
\newcommand{\AttnVisualMean}{0.430}
\newcommand{\AttnVisualLo}{0.251}
\newcommand{\AttnVisualHi}{0.578}
\newcommand{\AttnModelDelta}{0.436}
\newcommand{\AttnModelLo}{0.215}
\newcommand{\AttnModelHi}{0.646}
\newcommand{\AttnMiddleMean}{0.304}

\newcommand{\AttnCommonVisualMean}{0.430}
\newcommand{\AttnCommonVisualLo}{0.251}
\newcommand{\AttnCommonVisualHi}{0.578}

\begin{figure*}[!t]
\centering
\includegraphics[width=\linewidth]{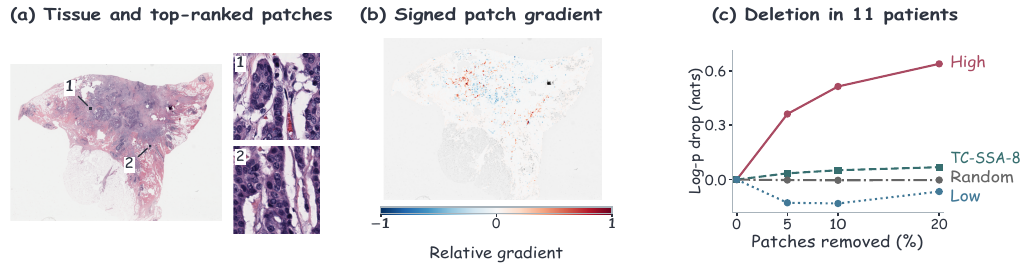}
\caption{\textbf{Testing patch gradients by deletion.} (a,b) Case 318, its two highest-ranked patches, and the signed \ours gradient scaled by its absolute maximum. (c) Mean log-probability drop for each model's original prediction on the same 11 patients. The dashed curve shows high-gradient deletion for TC-SSA-8. The other three show \ours. Random deletion averages five masks per patient.}
\label{fig:gradient_deletion}
\end{figure*}

\paragraph{Reads through the language model.}
In case 318 (\cref{fig:attention_capture}), \emph{subtype} gives slot~4 the largest mean attention weight (\cref{fig:attention_tokens}). We recorded 532 attention-block interventions across both models and tested every layer on 11 held-out breast-subtype slides. Blocking all visual reads lowers \ours's benchmark-answer log probability by \AttnCommonVisualMean{} nats on average. Layers 15--21 give the largest group means for \ours and TC-SSA-8 (0.304 and 0.080 nats). Layer~16 gives the largest mean single-layer drop (0.175 and 0.038 nats, \cref{fig:vet_depth}c). Report attributes are decodable by layer~2.

\paragraph{Slot information and answer influence.}
On 780 questions from 530 source patients (\cref{tab:vet_focus}), the within-attribute correlation between slot probe AUC and substitution effect averages $\rho=\cdErho$ for \ours-8 and $\cdBrho$ for TC-SSA-8. The three highest-AUC slots have a larger mean effect than the other five for \cdEngt/\cdNattr{} attributes in \ours-8 and \cdBngt/\cdNattr{} in TC-SSA-8. Shuffling attribute names while keeping slot identities fixed gives 0.480 and 0.161, respectively. Much of the agreement in \ours thus reflects slots that matter across attributes (\cref{fig:vet_specificity}). Attention ranks correlate more weakly with substitution effects on 150 questions ($\cdErhoAttn$ and $\cdBrhoAttn$).

\begin{table}[t]
\centering\small
\caption{\textbf{Tissue routed to each token on 1,425 slides.} Share is the mean fraction of routing weight (\%) by CONCH class. Enrichment subtracts same-slide patch prevalence. Both include uncertain patches.}
\label{tab:slot_tissue_routing}
\begin{tabular*}{\linewidth}{@{\extracolsep{\fill}}clrlr@{}}
\toprule
Token & \multicolumn{2}{c}{Largest tissue share} & \multicolumn{2}{c}{Largest enrichment} \\
\cmidrule(lr){2-3}\cmidrule(l){4-5}
 & Tissue & \% & Tissue & pp \\
\midrule
1 & Tumour & 25.1 & Adipose & +6.5\\
2 & Tumour & 23.1 & Muscle & +5.2\\
3 & Tumour & 28.4 & Necrosis & +1.3\\
4 & Tumour & \TableBest{72.8} & Tumour & \TableBest{+43.9}\\
5 & Stroma & 24.3 & Stroma & \TableSecond{+10.4}\\
6 & Tumour & \TableSecond{29.3} & Mucin & +4.6\\
7 & Stroma & 19.2 & Stroma & +5.3\\
8 & Muscle & 18.1 & Muscle & +8.3\\
\bottomrule
\end{tabular*}
\end{table}

\paragraph{Tissue behind the slots.}
Across 1,425 slides (\cref{tab:slot_tissue_routing}), tumour-labelled patches make up 28.9\% of the input and receive 72.8\% of slot~4's routing weight. Slots~5 and~7 favour stroma. Tumour has the largest share in slot~6, but mucin shows the greatest enrichment relative to its prevalence on the slide. For pT, slot~1 leads on both probe AUC and substitution effect. For pN, slot~6 has the strongest probe and slot~4 the largest effect (\cref{fig:vet_specificity}).

\begin{figure}[!t]
\centering
\includegraphics[width=\linewidth]{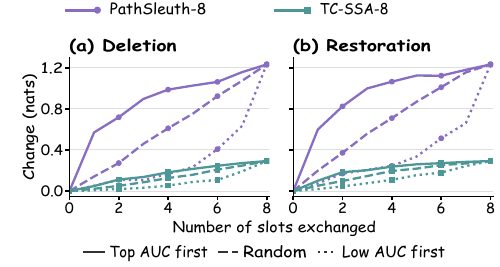}
\caption{\textbf{How many slots carry the answer?} Across 780 questions, (a) gives the drop in correct-answer log probability as donor slots replace original slots; (b) gives the gain as original slots return. Both patients are held out when ranking slots.}
\label{fig:vet_cumulative}
\end{figure}

Gradient deletion (\cref{fig:gradient_deletion}) tests which patches affect the answer. On the 11 indexed slides we rank 150,950 patches by the gradient of the predicted answer's log probability with respect to each patch's inclusion weight. Deleting the top 10\% per slide and recomputing all eight slots lowers that log probability by 0.513 nats on average for \ours and 0.052 for TC-SSA-8. Random deletion gives $-0.004$ nats for \ours. High-gradient deletion beats random deletion on all 11 slides, including one initially misclassified by \ours (\cref{tab:patch_cases}).

\paragraph{Restoring the original evidence.}
In \cref{fig:vet_cumulative}, all eight slots initially come from the donor. Restoring the highest-AUC slot recovers 0.60 nats of correct-answer log probability in \ours-8. Restoring the next two adds 0.40 nats (patient-bootstrap 95\% CI [0.31, 0.49]), bringing recovery to 1.00 of the 1.23 nats lost on a full swap (81.0\%). Three random slots recover 0.56 nats. TC-SSA-8 recovers 0.20 of 0.29 nats with its top three. Starting from the original slide, replacing the top three costs 0.90 nats in \ours, compared with 0.11 for the bottom three. Probe training excludes both patients' folds. The readers share donors and random slot orders.

\begin{table}[t]\centering\scriptsize\setlength{\tabcolsep}{1.5pt}
\caption{\textbf{Supervision on the full instruction set.} VR and all orders use SlideBench-VR. Pair scores intact pairs and Donor scores 1,124 regional exchanges (\%). The last row extends both training stages. Arrows show point differences from that row.}
\label{tab:supervision_recipes}
\begin{tabular*}{\linewidth}{@{\extracolsep{\fill}}lrrrr@{}}\toprule
Training recipe & VR & All orders & Pair & Donor \\\midrule
\rowcolor{white}
\multicolumn{5}{l}{\emph{One instruction epoch}} \\
\rowcolor{white}
Frozen visual modules & 41.6\TableDown{8.0} & 18.7\TableDown{12.3} & 19.7\TableDown{13.6} & 8.9\TableDown{23.1}\\
\rowcolor{PaperTableStripe}
Trainable visual modules & 42.5\TableDown{7.1} & 20.4\TableDown{10.6} & 19.6\TableDown{13.7} & 9.5\TableDown{22.5}\\
\rowcolor{white}
+ attribute heads & 47.3\TableDown{2.3} & 25.3\TableDown{5.7} & 24.7\TableDown{8.6} & 10.7\TableDown{21.3}\\
\rowcolor{PaperTableStripe}
+ whole-slide targets & 44.1\TableDown{5.5} & 17.9\TableDown{13.1} & 28.7\TableDown{4.6} & 10.1\TableDown{21.9}\\
\rowcolor{white}
+ heads and slide targets & \TableSecond{48.6}\TableDown{1.0} & \TableSecond{26.8}\TableDown{4.2} & \TableSecond{32.1}\TableDown{1.2} & \TableSecond{15.6}\TableDown{16.4}\\
\midrule
\rowcolor{white}
\multicolumn{5}{l}{\emph{Two instruction epochs}} \\
\rowcolor{PaperTableResult}
\ours-8 (full recipe) & \TableBest{49.6} & \TableBest{31.0} & \TableBest{33.3} & \TableBest{32.0}\\
\bottomrule\end{tabular*}\end{table}

\subsection{What changes when visual evidence guides training?}
\label{sec:ablation_transfer}
With one instruction epoch (\cref{tab:supervision_recipes}), visual-module and attribute-head training gives 24.7\% paired correctness. Whole-slide targets give 28.7\% with trainable visual modules and 32.1\% when combined with the heads. The two-epoch \ours-8 reaches 33.3\% on intact pairs and 32.0\% on regional exchanges, with test-set slot alignment $\rho=0.531$. Unfreezing the visual modules alone barely changes paired correctness. The heads must distinguish a report attribute across patients. Each paired question ties that attribute to an answer choice. Using both gives the strongest one-epoch result on intact pairs. The larger gain on regional exchanges in the final row also includes longer training.

A separate four-arm study holds training views and update counts fixed (\cref{tab:g2_matched}). Adding paired cross-entropy raises validation paired correctness by \GtwoGainDelta{} points (95\% CI [\GtwoGainLo, \GtwoGainHi]). This arm leads on 1,599 same-question validation pairs (\GtwoPairedPair\%) and slot alignment ($\rho=0.526$). The full-loss arm leads on joint donor following (\GtwoFullTr\%, versus \GtwoAuxTr\% with original-answer and attribute supervision). Best-slot AUC spans 0.762--0.766 across the four arms. On an intact pair, surrounding tissue may support the same diagnosis as the tumour. In a regional exchange, that tissue stays in place as the tumour donor changes. Scoring both exchange directions tests whether the prediction follows each donor in the other patient's tissue context.

\paragraph{Domain adaptation.}
\ours-SC combines SlideChat reader weights, option permutations and 25k replay examples. VR accuracy rises from 49.6\% to 54.0\% and zero-shot WSI-VQA* from 52.0\% to 62.7\% (\cref{tab:auxiliary}). Intact-pair accuracy falls from 33.3\% to 30.1\% and donor following from 32.0\% to 26.7\%. The paired gap reaches 7.33 points in training cancers absent from SlideInstruct (\cref{tab:generalization}). Both readers use eight slots, so the difference persists at a fixed visual budget.

\begin{table}[t]
\centering\small
\setlength{\tabcolsep}{2pt}
\caption{\textbf{Transfer and WSI-VQA fine-tuning.} Mean and all-orders accuracy (\%) on BCNB (7,274 questions) and WSI-VQA* (75/79 questions from 16/17 patients).}
\label{tab:auxiliary}
\begin{tabular*}{\linewidth}{@{\extracolsep{\fill}}l cc cc@{}}
\toprule
& \multicolumn{2}{c}{BCNB} & \multicolumn{2}{c}{WSI-VQA*} \\
\cmidrule(lr){2-3}\cmidrule(lr){4-5}
Model & Mean & All & Mean & All \\
\midrule
\rowcolor{white}
\multicolumn{5}{l}{\emph{Zero-shot transfer}} \\
\rowcolor{PaperTableStripe}
TC-SSA-32 & 52.6 & 42.9 & \TableSecond{56.0} & \TableSecond{34.7}\\
\rowcolor{white}
\ours-8 (5/1 epochs) & 58.4 & 52.4 & 52.0 & \TableSecond{34.7}\\
\rowcolor{PaperTableStripe}
LoC-Path + \ours & \TableBest{61.4} & \TableBest{57.1} & 51.7 & 30.7\\
\rowcolor{white}
\ours-SC & \TableSecond{59.6} & \TableSecond{52.8} & \TableBest{62.7} & \TableBest{52.0}\\
\midrule
\rowcolor{white}
\multicolumn{5}{l}{\emph{WSI-VQA fine-tuning}} \\
\rowcolor{PaperTableStripe}
\multicolumn{3}{l}{\ours-8, before tuning (15/2 epochs)} & \TableSecond{48.0} & \TableSecond{30.7}\\
\rowcolor{PaperTableResult}
\multicolumn{3}{l}{\ours-8, after tuning} & \TableBest{60.0} & \TableBest{46.7}\\
\bottomrule
\end{tabular*}
\end{table}

\paragraph{Which errors change?}
Among 3,994 questions, TC-SSA-8 makes 73 errors where probes support the true label but a full swap changes its log probability by at most 0.10 nat. \ours-8 corrects 47 and makes 18 errors of this kind. High-support errors with larger changes rise from 188 to 285. In 239 of its 1,408 errors, no probe gives the true label more than uniform probability (\cref{tab:vet_errors}). When a probe favours the report label, the swap can still move that label's score even if another option wins. High-support, small-swap errors remain more frequent in TC-SSA-8 across all nine combinations of support and effect thresholds (\cref{sec:vet_errors}).

\paragraph{Transfer across datasets.}
LoC-Path + \ours trails \ours-8 by 0.6 points on intact pairs. On BCNB biopsies~\citep{Xu2021BCNB} it leads by 4.7 points in all-orders accuracy, and on the 109 questions missed by all text judges it leads by 3.9 (\cref{tab:auxiliary,tab:visual_must}). Fine-tuning on WSI-VQA~\citep{Chen2024WSIVQA} raises \ours-8 accuracy from \auxWSIBeforeMean\% to \auxWSIAfterMean\% on 75 held-out questions. These rows use the 15/2-epoch starting model and the same patients. The zero-shot eight-slot row uses the 5/1 schedule. We select the fine-tuning epoch on separate validation patients. The share answered correctly in every option order rises from 30.7\% to 46.7\% (\cref{tab:auxiliary}). The patient-bootstrap 95\% interval for the all-orders gain is [6.4, 26.3] percentage points.

\section{Conclusion}
\label{sec:conclusion}
Eight question-independent slots retain report attributes that a reader can overlook. After evidence-anchored training, the correlation between slot probe rank and answer influence rises from 0.23 to 0.53. \ours-8 also corrects 47 of 73 baseline errors where a probe supported the label but a slide swap barely affected its score. Patch deletion traces this dependence to tissue on 11 slides. The 96-token LoC-Path variant scores higher on BCNB and on questions missed by all text judges. Its selector and training schedule both differ from the eight-slot transfer model (\cref{tab:auxiliary,tab:visual_must}).

\label{document:main_end}
{
  \small
  \setlength{\bibsep}{1pt plus 0.5pt minus 0.5pt}
  \bibliographystyle{ieeenat_fullname}
  \bibliography{main}
}
\PaperSuppModetrue
\clubpenalty=10000
\widowpenalty=10000
\displaywidowpenalty=10000
\clearpage
\maketitlesupplementary
\phantomsection
\label{document:supplement_start}
\appendix
\counterwithin{table}{section}
\renewcommand{\thetable}{\thesection\arabic{table}}
\setcounter{figure}{0}
\renewcommand{\thefigure}{S\arabic{figure}}
\renewcommand{\theHfigure}{supp.S\arabic{figure}}
\suppressfloats[t]

\section*{Overview and Protocol Guarantees}
This supplement provides the dataset construction details, training controls, visual evidence tracing analyses, supervision ablations, and external transfer evaluations supporting the main paper (\cref{sec:app_bench,sec:app_impl,sec:app_results,sec:app_vet,sec:app_design,sec:app_locator,sec:app_weighting}). Before the section-by-section analyses and tables, we summarise four protocol boundaries that govern all empirical comparisons.

\textbf{1. Non-circular benchmark filtering and patient-level split hygiene.}
None of the evaluated vision-language models participates in constructing SlideBench-VR or SlideBench-Evidence (\cref{sec:app_vr,sec:app_data}). Evaluation patients remain strictly excluded from caption alignment, instruction tuning, WSI-VQA adaptation, probe folds, and donor pools. Alternative text-judge panels, blind-removal filters, cross-fitted text-only readers, and patient-level pair reweighting leave the main rankings unchanged (\cref{tab:vr_sensitivity,tab:certify,tab:patient_weight}).

\textbf{2. Matched-budget training controls and inference efficiency.}
Extending the frozen eight-slot TC-SSA-8 baseline to the full caption and instruction schedule leaves its main-table metrics unchanged, whereas adding attribute heads and relational objectives on identical data drives the gains in paired correctness and donor following (\cref{sec:app_impl,tab:ablation}). A strictly update-matched four-arm study isolates each supervision term under shared Stage-1 weights and forward views (\cref{tab:g2_matched}), while the eight-token interface cuts prefill compute by more than two orders of magnitude relative to full-slide reading (\cref{tab:efficiency}).

\textbf{3. Non-circular regional exchange and locator-independent transfer.}
Regional Transplant fixes recipient context and exchanged patch count across tumour, non-tumour, and same-label controls on identical slides so that donor following cannot arise from patch-count shifts or locator artefacts (\cref{sec:app_locator,tab:transplant_controls}). Independent TCGA pathologist estimates validate the tumour and necrosis locator across cancer types, while intact-pair accuracy, unseen-cancer generalisation, SLICEChat held-out parity, BCNB transfer, and WSI-VQA adaptation never invoke the locator at test time (\cref{tab:heldout,tab:generalization,tab:auxiliary,tab:wsivqa_transitions}).

\textbf{4. Causal consistency from patches to slots to reader layers.}
Visual Evidence Tracing pairs patient-stratified linear probes with slot substitutions, exact tile-coordinate verification, patch-gradient deletion, and layerwise attention blocking (\cref{sec:app_vet,tab:patch_cases,tab:attention_focus,tab:attention_inference}). Across attributes, probe-ranked slot restoration recovers most of the full-swap log-probability drop and substantially reduces errors where decodable visual evidence fails to move the baseline reader (\cref{fig:app_attribute_restore,tab:vet_errors}).

\section{Benchmarks and Label Provenance}
\label{sec:app_bench}
This section documents the construction, label provenance, and shortcut controls of our evaluation suites. \Cref{sec:app_vr} details the non-circular text-only filtering, report audit, and sensitivity checks behind SlideBench-VR, while \cref{sec:app_data} specifies the report-derived SlideBench-Evidence and SlideBench-EvidenceLLM cohorts and patient-disjoint text controls.

\subsection{SlideBench-VR}
\label{sec:app_vr}
\begin{table}[t]
\centering
\small
\setlength{\tabcolsep}{2.6pt}
\caption{\textbf{Sensitivity to text-only filtering.} Four-order accuracy and CircularEval under each filter. No-slide removal excludes items solved blind by SlideChat, TC-SSA-32 or the default reader.}
\label{tab:vr_sensitivity}
\resizebox{\linewidth}{!}{%
\begin{tabular}{lrccc}
\toprule
Subset definition & $n$ & SlideChat & \ours & LoC-Path + \ours \\
\midrule
\rowcolor{PaperTableStripe}
Default panel & 1,178 & 39.9 / 8.5 & \TableBest{49.6} / \TableBest{31.0} & \TableSecond{49.2} / \TableSecond{28.1}\\
\rowcolor{white}
At least two judges solve & 1,715 & 46.0 / 14.2 & \TableBest{53.9} / \TableBest{34.4} & \TableSecond{52.2} / \TableSecond{30.6}\\
\rowcolor{PaperTableStripe}
Trained judges only & 1,587 & 43.7 / 11.9 & \TableBest{51.7} / \TableBest{32.1} & \TableSecond{50.3} / \TableSecond{28.8}\\
\rowcolor{white}
General LLMs only & 1,727 & 53.0 / 23.9 & \TableBest{60.3} / \TableBest{43.0} & \TableSecond{58.4} / \TableSecond{38.5}\\
\rowcolor{PaperTableStripe}
Default + no-slide removal & 384 & 22.3 / 1.3 & \TableSecond{33.3} / \TableSecond{20.8} & \TableBest{37.6} / \TableBest{21.9}\\
\rowcolor{white}
VR-HE & 642 & 39.6 / 7.6 & \TableBest{55.0} / \TableBest{37.7} & \TableSecond{54.0} / \TableSecond{33.0}\\
\rowcolor{PaperTableStripe}
VR-HE + no-slide removal & 203 & 22.8 / 1.0 & \TableSecond{40.6} / \TableSecond{28.1} & \TableBest{46.2} / \TableBest{28.6}\\
\bottomrule
\end{tabular}}
\end{table}

A breast-subtype prompt can often be answered from wording priors even when the underlying diagnosis requires microscopic tissue inspection. Starting from the expanded 3,176-question SlideBench release with its published stems and options intact, we complement pathology knowledge benchmarks such as PathMMU~\citep{Sun2024PathMMU} and multiscale reasoning suites such as PathView-Bench and EviPathBench~\citep{Chen2026PathView,Liao2026EviPathBench}. Following image-free solvability protocols in MMStar and MIRAGE~\citep{Chen2024MMStar,Asadi2026Mirage}, we filter items solved consistently across cyclic option permutations by a fixed text-only panel, audit keys against TCGA reports, verify stability across alternative filters, and extend evaluation to contrastive patient pairs.

\paragraph{Text-only judges.}
Each judge evaluates every SlideBench question across all four cyclic option orders without visual tokens; a question is flagged as text-solved if any judge answers all four rotations correctly. The panel combines four fixed models: Qwen2.5-7B with LoRA trained without visual inputs on official SlideChat data, the same text-only architecture trained for three epochs on all SlideInstruct questions, zero-shot Qwen2.5-7B-Instruct, and Gemini 3.1 Flash-Lite given only the stem and options. No evaluated VLM participates in question selection, keeping benchmark filtering independent of the compared visual models.

\paragraph{Report audit and answer balance.}
An independent audit compares the full TCGA pathology report against each benchmark question and answer key to remove report-contradicted items. Excluding the union of text-solved and contradicted items produces the SlideBench-VR benchmark and its single-section H\&E-determinable subset (SlideBench-VR-HE), whose sample sizes appear in \cref{tab:vr_sensitivity}. Because the published SlideBench answer key is skewed across option letters, scoring all four cyclic orders places every target equally across letters and removes letter-bias shortcuts.

\paragraph{Filter sensitivity and visual-must subset.}
\Cref{tab:vr_sensitivity} tests alternative judge panels, majority-judge thresholds, MIRAGE-style removal of questions solved without a slide by SlideChat, TC-SSA-32, or the base reader, and the H\&E-determinable subset. \ours leads SlideChat in both four-order accuracy and order-consistent CircularEval under every filtering criterion; the margin does not depend on any single panel definition. The strictest visual-must (VM) subset retains items missed in all four rotations by every text judge and verified by the report audit, all lying inside SlideBench-VR. \Cref{tab:visual_must} compares full-release four-order accuracy, no-slide accuracy, same-cancer swap drop, option-B bias, and VM accuracy across models. For protocol transparency, \cref{tab:visual_must,tab:auxiliary} evaluate the saved step-2,000 transfer snapshot of LoC-Path + \ours, whereas \cref{tab:main,tab:main_detailed} evaluate its final step-2,044 checkpoint.

\begin{table}[t]
\centering
\small
\setlength{\tabcolsep}{3pt}
\caption{\textbf{Four-order accuracy with and without slides.} VM retains 109 questions missed by all text judges in every order. $\Delta_\mathrm{swap}$ is the same-cancer replacement drop.}
\label{tab:visual_must}
\resizebox{\linewidth}{!}{%
\begin{tabular}{lcccccc}
\toprule
Model & 4-ord. & None & $\Delta_\text{swap}$ & Pred.\ B & VM & VM none \\
\midrule
\rowcolor{PaperTableStripe}
SlideChat & 64.4 & 63.4 & 0.2 & 49 & 18.8 & \TableBest{16.1}\\
\rowcolor{white}
SLICEChat & 65.3 & 60.7 & 3.6 & 43 & 21.3 & 11.3\\
\rowcolor{PaperTableStripe}
HistoSelect & 65.1 & \TableSecond{64.6} & 0.0 & 50 & 7.1 & 5.7\\
\rowcolor{white}
LoC-Path + \ours & 67.7 & 63.0 & \TableSecond{5.4} & 43 & \TableBest{29.6} & 11.9\\
\rowcolor{PaperTableStripe}
TC-SSA-8 & 67.0 & 61.1 & 2.6 & 45 & 14.2 & 3.7\\
\rowcolor{PaperTableResult}
\ours & \TableSecond{69.6} & 63.2 & \TableBest{6.7} & 43 & 25.7 & 5.3\\
\rowcolor{PaperTableStripe}
\ours-SC & \TableBest{74.2} & \TableBest{69.3} & 5.0 & 39 & \TableSecond{27.1} & \TableSecond{14.4}\\
\bottomrule
\end{tabular}}
\end{table}

\paragraph{From question filtering to patient pairs.}
Text filtering removes wording shortcuts on individual prompts, while SlideBench-Evidence tests whether answers track patient-specific tissue evidence. Each eligible slide is paired with a second patient of the same cancer type whose pathology report records a different value for the queried attribute. Both correct credits a pair only when both original questions are answered accurately, so cancer-level priors cannot satisfy both sides; Keep measures post-swap accuracy under same-label replacement (\cref{tab:heldout}).

\subsection{Report-Derived Questions}
\label{sec:app_data}
\begin{table}[t]
\centering
\small
\setlength{\tabcolsep}{3pt}
\caption{\textbf{Text-only controls on report-derived questions.} Accuracy across 1,356 questions, 647 slides and 13 finding families. Swap uses another slide of the same cancer type.}
\label{tab:certify}
\resizebox{\linewidth}{!}{%
\begin{tabular}{lccc}
\toprule
Reader & real & swap & none \\
\midrule
\rowcolor{PaperTableStripe}
Uniform guess / majority label & \multicolumn{3}{c}{43.5 / 52.9}\\
\rowcolor{white}
Text-only: zero-shot Qwen2.5-7B & -- & -- & 46.5\\
\rowcolor{PaperTableStripe}
Text-only: SFT on SlideInstruct & -- & -- & 45.9\\
\rowcolor{white}
Text-only: SFT on synthesised (cross-fit) & -- & -- & 49.3 / 43.1\\
\midrule
\rowcolor{PaperTableStripe}
TC-SSA ($K{=}32$) & 49.6 & 47.5 & \TableSecond{47.9}\\
\rowcolor{white}
$K{=}8$ + rule-based evidence supervision & 54.7 & \TableBest{50.8} & \TableBest{48.7}\\
\rowcolor{PaperTableStripe}
\quad + report token labels & \TableSecond{57.0} & \TableSecond{48.7} & \TableSecond{47.9}\\
\midrule
\rowcolor{white}
\multicolumn{4}{l}{With report-derived instruction training} \\
\rowcolor{white}
LoC-Path + \ours & \TableBest{58.8} & 46.2 & 45.4\\
\bottomrule
\end{tabular}}
\end{table}

To evaluate visual grounding beyond static SlideBench templates, we derive two complementary question cohorts directly from TCGA pathology reports. Below we detail the deterministic 13-attribute pipeline for SlideBench-Evidence, the quote-verified H\&E morphology extraction across seen and unseen cancers, and the patient-disjoint text-only controls in \cref{tab:certify}.

\paragraph{Rule-based attributes (SlideBench-Evidence).}
Structured fields from TCGA pathology reports are mapped by deterministic, cancer-specific rules to canonical labels across the 13 attributes ordered in \cref{fig:vet_specificity}: histologic grade (1--3), glioma grade (2--4), binary grade (low versus high), breast and glioma histologic subtypes, pathologic T and N categories of the staged specimen, tumour necrosis, lymphovascular invasion, perineural invasion, and assay-reported ER, PR, and HER2 status. During training we cap each attribute's majority-to-minority ratio at 2:1 before template sampling. At evaluation, natural cohort prevalence is preserved under one shared stem per attribute with shuffled options, spanning 3,994 questions across 1,425 slides (with 60 balanced substitutions per attribute for tracing in \cref{fig:vet_specificity}).

\paragraph{H\&E finding families and unseen cancers.}
A second extraction pass targets 13 H\&E-assessable morphological families paired with verbatim report quotes, excluding stage, nodal status, margins, gross size, and immunohistochemistry so every item reflects visible tissue morphology. Nine families use closed vocabularies (differentiation, tumour necrosis, lymphovascular invasion, perineural invasion, mitotic activity, nuclear pleomorphism, inflammatory infiltrate, in situ component, and tumour border), and four use open vocabularies (histologic subtype, growth pattern, depth of invasion, and organ-specific grade) whose synonyms are grouped within cancer type and retained when present on at least five training slides. Verbatim report substrings match 99.6\% of benchmark quotes. Applying both extraction pipelines to public TCGA reports~\citep{Kefeli2024TCGAReports} outside SlideInstruct, with all test patients strictly excluded, builds the expanded training set $\mathcal{D}_{\mathrm{ev+}}$ alongside the SlideBench-EvidenceLLM seen-cancer and unseen-cancer evaluation cohorts (\cref{tab:certify,tab:generalization}).

\paragraph{Text-only controls.}
Whereas RUBi incorporates a question-only branch during training~\citep{Cadene2019RUBi}, \cref{tab:certify} uses zero-shot, SlideInstruct-tuned, and cross-fitted text-only readers strictly as evaluation controls. Readers without slide access stay below the majority-label prior on held-out slides even when fine-tuned directly on synthesised training questions; shared stems remove template shortcuts. In contrast, slide-conditioned models trained with report supervision rise clearly above the prior on real slides and drop when evaluated on same-cancer swaps or without slides.

\section{Training the Eight-Slot Model and Controls}
\label{sec:app_impl}
\begin{table*}[t]
\centering
\caption{Notation used in the method, tracing analysis and training algorithm.}
\label{tab:notation}
\small
\setlength{\tabcolsep}{6pt}
\begin{tabularx}{\textwidth}{lL}
\toprule
Symbol & Definition \\
\midrule
$s,\bar s,\tilde s$ & Original slide, different-label donor, and same-label donor for the queried attribute.\\
$a,k,i,r$ & Attribute, visual-token position, micro-batch row, and relation index, respectively.\\
$X_s,N_s,D$ & Patch-feature matrix, number of patches in slide $s$, and patch-feature dimension.\\
$E,g,K,H$ & Slide encoder, projector, token budget, and reader embedding dimension.\\
$V_s,v_{s,k},\bar v_s$ & Projected tokens, their $k$th vector, and their mean over $K$ positions.\\
$\theta,\pi_\theta(y\mid q,V)$ & Trainable model parameters and reader probability of answer $y$ given question $q$ and tokens $V$.\\
$y_{s,a},\mathcal{Y}_a$ & Report label for attribute $a$ on slide $s$, and candidate labels for that attribute.\\
$t_a,\sigma,q_{s,a}$ & Shared question stem, option permutation, and resulting multiple-choice question.\\
$p_V,\ell(y,V)$ & Reader distribution over valid options under fixed $q$, and $-\log p_V(y)$.\\
$X'_s,V'_s,\rho$ & Subsampled patch set, its projected tokens, and patch-retention probability.\\
$M_s,\neg M_s$ & Locator-selected patch indices and their complement within $\{1,\ldots,N_s\}$.\\
$\widehat X_{\bar s\to s}$ & Donor-region feature rows sampled to match the recipient region's patch count.\\
$X_{s\leftarrow\bar s},V_{s\leftarrow\bar s}$ & Regional-exchange input and its projected tokens; the left index is the recipient.\\
$V_s^{\,k\leftarrow\bar s},\Delta_k(q,s)$ & Single-token substitution and its signed drop in original-answer log probability.\\
$B,B_a,A_B,h_a$ & Micro-batch, rows labelled for $a$, nonempty attribute set, and attribute head.\\
$I_r,y_i^{(r)},V_i^{(r)},\bar\ell_r$ & Eligible rows, target, tokens, and eligible-row mean option loss for relation $r$.\\
$\mathcal{L}_{\mathrm{LM}},\mathcal{L}_{\mathrm{enc}}$ & Supervised-answer autoregressive loss and base encoder regularisation.\\
$\mathcal{L}_{\mathrm{pre}},\mathcal{L}_{\mathrm{alt}},\mathcal{L}_{\mathrm{keep}},\mathcal{L}_{\mathrm{tr}}$ & Preserve, Alter, Keep, and Transplant, the regional form of Alter.\\
$\mathcal{L}_{\mathrm{lab}},\mathcal{L}_{\mathrm{rel}},\mathcal{L}$ & Attribute loss, sum of eligible-row relation means, and total training objective.\\
$\mu,\lambda,\gamma,\beta$ & Weights for Preserve LM, encoder regularisation, attributes, and relations.\\
$\mathrm{sg},\mathrm{KL},\mathrm{CE}$ & Stop-gradient, Kullback-Leibler divergence, and categorical cross-entropy.\\
$p,u,v$ & Slide-independent option distribution and target probabilities in the paired bound (\cref{sec:app_design}).\\
$\mathcal{D}_{\mathrm{ev+}},r_{\mathrm{LoRA}},\alpha_{\mathrm{LoRA}}$ & Common instruction-training dataset, LoRA adapter rank, and LoRA scaling parameter.\\
\bottomrule
\end{tabularx}
\end{table*}

This section provides the notation (\cref{tab:notation}), training setup, and matched controls supporting our experiments. \Cref{sec:app_protocol} specifies the three-stage schedule, objective regularisation, and scoring rules; \cref{sec:app_baselines} documents released checkpoints and re-implemented visual modules; and \cref{sec:app_cost} details the prefill FLOPs, latency, and memory cost.

\subsection{Training and Evaluation Protocol}
\label{sec:app_protocol}
Training proceeds in three stages using the notation in \cref{tab:notation}. Stage~0 pretrains the MaskedSlotCompressor on SlideInstruct slides for tumour-type classification with $K=8$ slots, top-2 softmax routing, Gaussian gate noise $\sigma=0.1$, and encoder regularisation weight $0.1$. Stage~1 aligns the slot encoder and a two-layer GELU MLP projector (dimensions 512, 2,048, and 3,584) on 4,181 captions for 15 epochs at global batch size 288 (four workers, micro-batch 4, 18 accumulation steps), with peak learning rates $10^{-4}$ for the encoder and $10^{-3}$ for the projector, cosine decay after 10\% warmup, and seed 42.

Stage~2 fine-tunes on the instruction set $\mathcal{D}_{\mathrm{ev+}}$ for two epochs (2,044 optimizer updates) at global batch size 128 (four workers, micro-batch 2, 16 accumulation steps, seed 43). Qwen2.5-7B-Instruct uses rank-64 LoRA ($r_{\mathrm{LoRA}}=64$, $\alpha_{\mathrm{LoRA}}=128$) across attention and MLP projections at peak learning rate $2\times 10^{-4}$, while the encoder, projector, and LayerNorm-linear attribute heads $\{h_a\}$ use $10^{-4}$. Learning rates warm up linearly for 100 updates, hold constant through half the schedule, and decay by cosine to 10\% of peak; incomplete trailing batches are dropped so every step uses an exact global batch. Extending the frozen TC-SSA-8 control from the shorter 5/1-epoch schedule to this full 15/2-epoch schedule leaves all main-table metrics unchanged (\cref{tab:main,tab:ablation}). Longer instruction tuning alone does not elicit visual readout when slots stay frozen.

\paragraph{Objective coefficients and encoder regularisation.}
Encoder regularisation uses weight $0.2$ in Stage~1 and $\lambda=0.1$ in Stage~2, and attribute heads use $\gamma=1$ in both stages. In \cref{eq:pre}, the Preserve mask retains the first valid patch to guarantee a nonempty view and samples each remaining valid patch independently with probability $\rho=0.5$, using LM weight $\mu=0.5$ and unit KL weight on closed-form multiple-choice rows. Alter, Keep, and both Transplant directions each use $\beta=0.25$ in \cref{eq:total}, shared across \ours-8, \ours-SC, and the ablations in \cref{sec:app_design}. Donors are sampled once per epoch from distinct patients, preferring matching cancer types; Transplant replaces all located tumour patches with an equal-count sample from the donor tumour mask. Let $P_k$ be the mean softmax gate probability of slot $k$ over valid patches, $f_k$ the fraction whose top-2 slots include $k$, and $z_j \in \mathbb{R}^K$ the gate logits on valid patch $j$. The gate entropy $H(P)$ and encoder regulariser $\mathcal{L}_{\mathrm{enc}}$ in \cref{eq:total} are defined as
\begin{equation*}
\begin{aligned}
H(P) &= -\sum_{k=1}^{K} P_k \log P_k,\\
\mathcal{L}_{\mathrm{enc}} &= K\sum_{k=1}^{K} P_k f_k + \frac{1}{2}\!\left(1 - \frac{H(P)}{\log K}\right) \\
&\quad + 10^{-4}\!\left(\operatorname{mean}_{j}\log\sum_{k=1}^{K} e^{z_{jk}}\right)^{\!2}\!.
\end{aligned}
\end{equation*}
Padding patches are excluded from routing and $\mathcal{L}_{\mathrm{enc}}$, and $\mathcal{L}_{\mathrm{LM}}$ supervises only assistant tokens after the prompt prefix.

\paragraph{Loss reductions and evaluation scoring.}
\label{sec:app_reduction}
Within each worker's per-device micro-batch, each active relation $r$ in the set of Alter, Keep, and bidirectional Transplant views averages option-level cross-entropy over its eligible rows $I_r$, and the relational objective sums the active relation means without dividing by the number of active relations:
\begin{equation*}
\begin{aligned}
\overline{\ell}_r &=
\begin{cases}
|I_r|^{-1}\sum_{i\in I_r}\ell\!\left(y_i^{(r)}, V_i^{(r)}\right), & |I_r| > 0,\\
0, & |I_r| = 0,
\end{cases}\\
\mathcal{L}_{\mathrm{rel}} &= \sum_{r \in \{\mathrm{alt},\,\mathrm{keep},\,\mathrm{tr}{\to},\,\mathrm{tr}{\leftarrow}\}} \overline{\ell}_r.
\end{aligned}
\end{equation*}
Preserve KL averages eligible multiple-choice rows with only the teacher distribution detached, Original and Preserve LM terms each average supervised answer tokens, and $\mathcal{L}_{\mathrm{lab}}$ averages labelled original rows per attribute before averaging nonempty attributes (\cref{eq:lab}). Gradients are scaled by $m=16$ local accumulation steps, averaged across four workers, and clipped to global $\ell_2$ norm $1.0$. At evaluation, multiple-choice answers are scored from fp32 log-softmax option-letter probabilities at the final prompt position, verified against deterministic greedy decoding with SlideChat's prompt. Same-cancer swaps assign a fixed random slide from another patient in the same TCGA project; paired bootstrap intervals use 1,000 resamples for broad comparisons and 10,000 resamples for focused eight-slot, microscopy, and matched-supervision contrasts.

\subsection{Compared Methods}
\label{sec:app_baselines}
\label{sec:app_notcompared}
Released pathology VLMs use their published weights and prompt formats, while re-implemented visual compressors share CONCH v1 features, Qwen2.5-7B-Instruct with rank-64 LoRA, and $\mathcal{D}_{\mathrm{ev+}}$ data (five caption epochs and one instruction epoch at native token budgets).

\paragraph{Released checkpoints.}
SlideChat~\citep{Chen2025SlideChat} applies its stage-2 LongNet encoder, projector, and Qwen reader to up to 38,400 CONCH patches per slide. SLICEChat~\citep{Bozkurt2026SLICEChat} runs its SlideInstruct checkpoint on $20\times$ 512-px CONCH-v1.5 features from TRIDENT~\citep{Zhang2025TRIDENT}, pruning slide patches down to a few hundred tokens on covered questions; greedy decoding matches option-letter scoring on checked items and reproduces published benchmark accuracy closely. HistoSelect~\citep{Huang2026HistoSelect} attaches its released LongNet, projector, tissue-prompt selectors, and LoRA adapters ($r_{\mathrm{LoRA}}=64$, $\alpha_{\mathrm{LoRA}}=16$) to Qwen2.5-7B-Instruct, scoring up to 10,000 subsampled CONCH features via its 13-class prompt ensemble and matching published accuracy under logit scoring.

\paragraph{Re-implemented visual modules.}
Because public weights were unavailable for TCP-LLaVA, LoC-Path, and SparseLearn/PathSelect, we re-implemented their visual interfaces on the shared backbone. \emph{TCP-LLaVA}~\citep{Lyu2025TCPLLaVA} compresses up to 10,000 patches and question tokens into 100 tokens through two reader-initialised cross-attention layers trained at learning rate $2\times 10^{-5}$ with LoRA reader adaptation. \emph{LoC-Path}~\citep{Hu2025LoCPath} merges four raster-adjacent patches, applies a two-layer Perceiver with 256 latents, and gates the top 96 question-scored latents into the prefix interface. \emph{SparseLearn/PathSelect}~\citep{Chen2026SparseLearn,Chen2026PathSelect} selects 32 CONCH patches via sequential straight-through Gumbel-Softmax ($\tau=0.1$) during training and hard top-$K$ at test time before a residual MLP. \emph{Mean pooling} projects a single mean patch vector, and \emph{TC-SSA}~\citep{Chen2026TCSSA} uses Stage-0 slot pretraining and caption alignment with frozen visual slots during instruction tuning.

\subsection{Computational Cost Accounting}
\label{sec:app_cost}
\Cref{tab:efficiency} compares visual token budgets, total LLM sequence lengths, prefill TFLOPs (covering visual compression, all 28 Qwen2.5-7B layers with four KV heads, and the option-letter head), cached-feature latency, and peak VRAM on one RTX~4090 over 57 benchmark slides after three warm-up passes. Benchmark slides average roughly sixteen thousand 224-px CONCH patches (and roughly three and a half thousand 512-px patches for SLICEChat, whose upper-bound FLOPs include unpruned encoder passes; HistoSelect excludes external tile-to-tissue encoding). Collapsing an entire slide into eight visual tokens shifts inference cost away from long-sequence LLM prefill, cutting prefill compute by more than two orders of magnitude relative to SlideChat at low single-GPU memory even under profiling.

\begin{table}[t]
\centering
\small
\setlength{\tabcolsep}{3pt}
\caption{\textbf{Cost per question on 1,292 SlideBench slides.} Latency runs from patch features to answer scores on one RTX~4090 without cache reuse. Memory is peak allocation.}
\label{tab:efficiency}
\resizebox{\linewidth}{!}{%
\begin{tabular}{lrrrrr}
\toprule
Method & tokens & LLM length & TFLOPs & latency (s) & memory (GB) \\
\midrule
\rowcolor{PaperTableStripe}
SlideChat$^\dagger$ & 15.9k & 15,964 & 279 & 2.073 & 18.8\\
\rowcolor{white}
SLICEChat$^\dagger$ & 705 & 799 & $\approx$11.1 & 0.146 & \TableSecond{16.2}\\
\rowcolor{PaperTableStripe}
HistoSelect$^\dagger$ & 2.1k & 2,194 & 31.5 & 0.274 & 17.8\\
\rowcolor{white}
Mean pooling & \TableBest{1} & \TableBest{85} & \TableBest{1.1} & -- & --\\
\rowcolor{PaperTableStripe}
TC-SSA & 32 & 116 & 1.5 & \TableSecond{0.090} & \TableSecond{16.2}\\
\rowcolor{white}
TCP-LLaVA$^\ast$ & 100 & 184 & 2.8 & -- & --\\
\rowcolor{PaperTableStripe}
LoC-Path$^\ast$ & 96 & 180 & 2.4 & -- & --\\
\rowcolor{white}
SparseLearn / PathSelect$^\ast$ & 32 & 116 & 1.5 & -- & --\\
\rowcolor{PaperTableStripe}
LoC-Path + \ours & 96 & 180 & 2.4 & \TableBest{0.074} & \TableSecond{16.2}\\
\rowcolor{PaperTableResult}
\ours & \TableSecond{8} & \TableSecond{92} & \TableSecond{1.2} & \TableSecond{0.090} & \TableBest{16.0}\\
\bottomrule
\end{tabular}}
\end{table}

\section{Counterfactual Answer Evaluation}
\label{sec:app_results}
\begin{table*}[t]
\centering
\small
\setlength{\tabcolsep}{6pt}
\caption{\textbf{Extended answer and evidence diagnostics.} Accuracy averages four cyclic orders; All orders requires four correct rotations; $\Delta_\text{swap}$ is same-cancer replacement drop; Both correct scores intact pairs; wAUC is within-cancer AUC (bold best, underline second).}
\label{tab:main_detailed}
\begin{tabularx}{\textwidth}{L ccc ccc c}
\toprule
& \multicolumn{3}{c}{SlideBench-VR} & \multicolumn{3}{c}{SlideBench-Evidence} & TCGA \\
\cmidrule(lr){2-4}\cmidrule(lr){5-7}
Method & Accuracy & All orders & $\Delta_\text{swap}$ & Both correct & Accuracy & wAUC & Accuracy \\
\midrule
\rowcolor{PaperTableStripe}
Text-only reader & 36.1 & 10.6 & -- & 5.6 & 51.9 & -- & 63.8\\
\midrule
\rowcolor{white}
SlideChat$^\dagger$ & 39.9 & 8.5 & 0.2 & 13.7 & 40.6 & .543 & 64.4\\
\rowcolor{PaperTableStripe}
SLICEChat$^{\dagger\ddagger}$ & 44.9 & 17.8 & 6.7 & 20.1 & 54.0 & .646 & 65.3\\
\rowcolor{white}
HistoSelect$^{\dagger\|}$ & 38.4 & 11.9 & $-$0.1 & 5.2 & 46.6 & .510 & 65.1\\
\midrule
\rowcolor{PaperTableStripe}
Mean pooling & 38.4 & 11.7 & 2.8 & 12.9 & 54.8 & .611 & 64.9\\
\rowcolor{white}
TC-SSA-32 & 41.6 & 14.2 & 3.3 & 15.0 & 60.2 & .602 & 66.4\\
\rowcolor{PaperTableStripe}
TC-SSA-8 & 41.6 & 18.7 & 4.2 & 19.7 & 59.9 & .649 & 67.0\\
\rowcolor{white}
TCP-LLaVA$^{\ast\S}$ & 35.4 & 5.2 & 0.1 & 5.6 & 51.0 & .513 & 62.9\\
\rowcolor{PaperTableStripe}
LoC-Path$^\ast$ & 40.2 & 13.2 & 4.8 & 20.8 & 57.0 & .631 & 66.1\\
\rowcolor{white}
SparseLearn / PathSelect$^\ast$ & 41.9 & 18.4 & 4.9 & 19.2 & 59.2 & .623 & 67.4\\
\midrule
\rowcolor{PaperTableStripe}
LoC-Path + \ours & 49.2 & 28.1 & -- & \TableSecond{32.7} & 63.4 & \TableSecond{.686} & 67.1\\
\rowcolor{PaperTableResult}
\ours & \TableSecond{49.6} & \TableSecond{31.0} & \TableBest{10.0} & \TableBest{33.3} & \TableBest{64.7} & \TableBest{.696} & \TableSecond{69.6}\\
\midrule
\rowcolor{PaperTableStripe}
\ours-SC (E3)$^\diamond$ & \TableBest{54.0} & \TableBest{42.0} & \TableSecond{7.5} & 30.1 & \TableSecond{63.6} & .669 & \TableBest{74.2}\\
\bottomrule
\end{tabularx}
\par\smallskip
\begin{minipage}{\textwidth}\footnotesize
$^\dagger$Released weights and original training data; $^\ast$visual-module re-implementations. $^\ddagger$CONCH-v1.5, 512-px features, 96--97\% coverage (\cref{tab:heldout}). $^\S$At most 10,000 input patches; $^\|$mean question-dependent budget. $^\diamond$E3: released reader, permutation and 25k replay (\cref{sec:official_variant}). Other trained rows share $\mathcal{D}_\text{ev+}$ and CONCH features; training schedules differ. \ours: 2,004 patches per token. Protocols and costs: \cref{sec:app_baselines,sec:app_cost}.
\end{minipage}
\end{table*}

\begin{table}[t]
\centering
\small
\setlength{\tabcolsep}{2.6pt}
\caption{\textbf{Evaluation on slides held out from SLICEChat pretraining.} All rows cover 1,483 SlideBench questions (513 VR) and 1,982 Evidence questions (1,466 pairs).}
\label{tab:heldout}
\resizebox{\linewidth}{!}{%
\begin{tabular}{lr ccccccc}
\toprule
& & SB & \multicolumn{2}{c}{SlideBench-VR} & VR & Evid. & \multicolumn{2}{c}{Evidence} \\
\cmidrule(lr){3-3}\cmidrule(lr){4-5}\cmidrule(lr){6-6}\cmidrule(lr){7-7}\cmidrule(lr){8-9}
Method & tokens & 4-ord. & 4-ord. & Circ. & $\Delta_\text{swap}$ & Pair & acc. & wAUC \\
\midrule
\rowcolor{PaperTableStripe}
SlideChat$^\dagger$ & 15.9k & 64.4 & 37.9 & 5.8 & 0.2 & 14.9 & 42.1 & .554\\
\rowcolor{white}
SLICEChat$^\dagger$ & 705 & \TableSecond{73.5} & \TableBest{55.0} & 28.7 & \TableSecond{12.1} & 24.7 & 58.8 & .647\\
\rowcolor{PaperTableStripe}
TC-SSA & 32 & 71.7 & 48.3 & 21.6 & 4.6 & 17.6 & \TableSecond{64.5} & .620\\
\rowcolor{PaperTableResult}
\ours & 8 & 72.8 & \TableSecond{54.9} & \TableBest{41.3} & \TableBest{12.9} & \TableSecond{35.3} & \TableBest{65.5} & \TableSecond{.710}\\
\rowcolor{PaperTableStripe}
LoC-Path + \ours & 96 & 69.3 & 52.4 & \TableSecond{36.8} & -- & 31.0 & 64.3 & .690\\
\rowcolor{white}
TC-SSA, CONCH-v1.5 & 8 & 71.9 & 48.1 & 25.7 & 3.7 & 18.8 & 63.3 & .640\\
\rowcolor{PaperTableStripe}
\ours, CONCH-v1.5 & 8 & \TableBest{73.9} & 54.8 & 36.5 & \TableSecond{12.1} & \TableBest{37.0} & 64.2 & \TableBest{.745}\\
\bottomrule
\end{tabular}}
\end{table}

\Cref{tab:main_detailed} extends the main benchmark table with same-cancer slide-swap drops ($\Delta_{\mathrm{swap}}$), within-cancer ROC-AUC (wAUC), and full four-order SlideBench TCGA accuracy alongside the compute profiles in \cref{tab:efficiency}. Across all eight-token and re-implemented visual baselines, \ours-8 delivers the largest accuracy drop under same-cancer slide replacement, the highest paired correctness, and the strongest within-cancer discrimination. This pattern shows that evidence-anchored training strengthens patient-specific visual sensitivity even at an eight-token bottleneck, whereas combining EvA with a pathology-pretrained reader in \ours-SC achieves the highest SlideBench-VR accuracy, CircularEval consistency, and overall TCGA accuracy.

\paragraph{Parity on slides held out from SLICEChat pretraining.}
SLICEChat's visual backbone was self-supervised on unlabeled TCGA slides after excluding the initial SlideBench v1 release and WSI-Bench test cohorts, leaving partial slide overlap with the expanded release. To remove that confound, \cref{tab:heldout} restricts evaluation strictly to slides excluded from SLICEChat pretraining across both SlideBench-VR and SlideBench-Evidence. Even with two orders of magnitude fewer visual tokens than SLICEChat, \ours-8 matches its four-order VR accuracy and establishes clear margins in order-consistent CircularEval and contrastive paired correctness. Training \ours-8 on CONCH-v1.5 features over the matching subset raises paired correctness and within-cancer discrimination further; the advantage persists under a matched visual encoder.

\subsection{Microscopy and Report-Derived Morphology}
\label{sec:app_microscopy}
\begin{table}[!tb]
\centering
\small
\setlength{\tabcolsep}{3pt}
\caption{\textbf{Four-order accuracy on 227 VR microscopy questions.} Swap uses another slide of the same cancer type. None removes the slide. Historical variants use different training schedules.}
\label{tab:microscopy}
\resizebox{\linewidth}{!}{%
\begin{tabular}{l ccc cc}
\toprule
& \multicolumn{3}{c}{VR microscopy} & \multicolumn{2}{c}{Histopath.\ changes} \\
\cmidrule(lr){2-4}\cmidrule(lr){5-6}
Model & own & swap & none & own & swap \\
\midrule
\rowcolor{PaperTableStripe}
SlideChat$^\dagger$ & \TableSecond{50.2} & \TableSecond{50.3} & \TableBest{47.6} & \TableSecond{57.0} & \TableSecond{57.0}\\
\rowcolor{white}
SLICEChat$^{\dagger\ddagger}$ & 43.2 & 41.6 & 39.7 & 50.0 & 47.8\\
\rowcolor{PaperTableStripe}
HistoSelect$^\dagger$ & 39.1 & 39.3 & 38.5 & 48.4 & 48.4\\
\rowcolor{white}
LoC-Path + \ours & 43.8 & 41.9 & 38.4 & 48.8 & 43.0\\
\rowcolor{PaperTableStripe}
LoC-Path$^\ast$ & 40.1 & 39.2 & 34.5 & 49.2 & 48.8\\
\rowcolor{white}
SparseLearn / PathSelect$^\ast$ & 38.9 & 40.1 & 37.0 & 38.1 & 41.0\\
\rowcolor{PaperTableStripe}
TC-SSA-8 & 38.1 & 36.7 & 32.8 & 40.2 & 38.9\\
\rowcolor{PaperTableResult}
\ours-8 & 45.9 & 42.4 & 38.2 & 47.5 & 47.5\\
\rowcolor{PaperTableStripe}
\ours-SC (E3) & \TableBest{51.4} & \TableBest{51.7} & \TableSecond{47.4} & \TableBest{59.4} & \TableBest{63.1}\\
\bottomrule
\end{tabular}}
\end{table}

\begin{table}[t]\centering\small
\setlength{\tabcolsep}{3pt}
\caption{\textbf{Microscopy accuracy by question category.} Each score averages four cyclic orders on the original slide ($N$ questions).}
\label{tab:microscopy_categories}
\resizebox{\linewidth}{!}{%
\begin{tabular}{lrrrrr}\toprule
Category & $N$ & TC-SSA-8 & \ours & \ours-SC & LoC-Path + \ours \\ \midrule
\rowcolor{PaperTableStripe}
Cytomorphology & 23 & \TableSecond{51.1} & 46.7 & \TableBest{58.7} & 45.7\\
\rowcolor{white}
Histopathological changes & 61 & 40.2 & 47.5 & \TableBest{59.4} & \TableSecond{48.8}\\
\rowcolor{PaperTableStripe}
Tissue architecture & 95 & 36.3 & \TableSecond{46.8} & \TableBest{51.6} & 42.4\\
\rowcolor{white}
Tumour characteristics & 48 & 32.8 & \TableBest{41.7} & 37.5 & \TableSecond{39.6}\\
\bottomrule\end{tabular}}
\end{table}

\Cref{tab:microscopy,tab:microscopy_categories} break down the SlideBench-VR microscopy questions to separate visual recognition from domain-language priors. At eight visual tokens, \ours-8 improves clearly over frozen TC-SSA-8 on overall microscopy accuracy and histopathological changes, and leads all compared models on tumour characteristics. Incorporating pathology-pretrained reader weights in \ours-SC lifts microscopy and histopathological-change accuracy further, accompanied by higher no-slide accuracy that reflects stronger domain-adapted language priors (\cref{sec:official_variant}).

\paragraph{Slide dependence on morphology and unseen cancers.}
Two additional analyses test whether predictions rely on patient-specific morphology when prompt wording is fixed or cancer types are unfamiliar. First, on the ten-family fixed-stem morphology subset drawn from the 13-family SlideBench-EvidenceLLM cohort (whose text-only controls appear in \cref{tab:certify}), \ours-8 shows a much steeper accuracy drop under same-cancer slide replacement than either TC-SSA-8 or SlideChat because its predictions track patient-specific tissue features. Second, on contrastive pairs drawn from cancer types absent from SlideInstruct (\cref{tab:generalization}), \ours-8 achieves the highest real-slide accuracy and paired correctness, ahead of both frozen TC-SSA-8 and \ours-SC. Because domain-specific language priors transfer less readily to unfamiliar malignancies, the wider paired-correctness margin on unseen cancers traces directly to patient-specific visual grounding.
\begin{table}[t]\centering\small
\setlength{\tabcolsep}{3pt}
\caption{\textbf{Cancers absent from SlideInstruct.} These cancers occur in $\mathcal{D}_\mathrm{ev+}$. Pair scores joint original correctness ($N$ pairs).}
\label{tab:generalization}
\resizebox{\linewidth}{!}{%
\begin{tabular}{llrrrrr}
\toprule
Set & Model & Real & Swap & Blind & Pair & $N$ \\ \midrule
\rowcolor{PaperTableResult}
U & \ours & \TableBest{61.5} & 48.1 & 44.2 & \TableBest{35.6} & 1105\\
\rowcolor{white}
U & Earlier reference & \TableSecond{59.9} & \TableSecond{48.2} & 45.5 & 27.6 & 1105\\
\rowcolor{PaperTableStripe}
U & Permutation only & 59.7 & 47.3 & 44.7 & \TableSecond{31.0} & 1105\\
\rowcolor{white}
U & Replay only & 58.8 & 47.0 & 44.6 & 27.6 & 1105\\
\rowcolor{PaperTableStripe}
U & \ours-SC & 58.0 & 46.5 & \TableSecond{46.6} & 28.2 & 1105\\
\rowcolor{white}
U & LoC-Path + \ours & 59.6 & \TableBest{49.2} & 45.8 & \TableSecond{31.0} & 1105\\
\rowcolor{PaperTableStripe}
U & TC-SSA-8 & 55.8 & \TableSecond{48.2} & \TableBest{47.6} & 19.3 & 1105\\
\bottomrule
\end{tabular}}
\end{table}

\section{Visual Evidence Tracing}
\label{sec:app_vet}
\label{sec:causal}
\begin{table}[t]
\centering
\small
\setlength{\tabcolsep}{2.5pt}
\caption{\textbf{Token attribution on 780 Evidence questions.} $\Delta$ is the drop in correct-answer log probability (nats). Attention and gradient correlations use 150 questions.}
\label{tab:causal}
\resizebox{\linewidth}{!}{%
\begin{tabular}{lcc}
\toprule
 & TC-SSA-8 & \ours-8 \\
Training data & $\mathcal{D}_\text{ev+}$ & $\mathcal{D}_\text{ev+}$ \\
\midrule
\rowcolor{PaperTableStripe}
Visual attention (uniform) & 4.4\% (13.6\%) & 7.3\% (13.6\%)\\
\rowcolor{white}
Slide swap: median / mean & \TableSecond{0.078} / \TableSecond{0.293} & \TableBest{0.722} / \TableBest{1.234}\\
\rowcolor{PaperTableStripe}
Slide swap: answer flips & 34.7\% & 59.2\%\\
\rowcolor{white}
Effective tokens (of $K$) & 3.8 / 8 & 3.3 / 8\\
\rowcolor{PaperTableStripe}
Information--effect correlation & \TableSecond{0.23} & \TableBest{0.53}\\
\rowcolor{white}
Top-three / other (count) & \TableSecond{1.12} (7/13) & \TableBest{4.44} (13/13)\\
\rowcolor{PaperTableStripe}
Tumour--effect correlation$^\ast$ & \TableSecond{0.24} & \TableBest{0.52}\\
\midrule
\rowcolor{white}
Attention vs.\ effect ($\rho$) & \TableSecond{0.07} & \TableBest{0.31}\\
\rowcolor{PaperTableStripe}
Grad.$\times$input vs.\ effect ($\rho$) & \TableBest{0.50} & \TableSecond{0.48}\\
\bottomrule
\end{tabular}}
\end{table}

This section traces how slide evidence moves from input patches through compressed visual slots to reader predictions. We pair cohort-wide probing, routing verification, and text-to-vision edge blocking (\cref{sec:app_slot_decode}) with single-slide substitutions (\cref{sec:visual_cases}), patch-gradient deletion (\cref{sec:patch_attribution}), layerwise attention knockouts (\cref{sec:attention_capture}), and per-attribute depth and error analyses (\cref{sec:vet_extended,sec:vet_errors}).

\subsection{Slot Decodability and Answer Influence}
\label{sec:app_slot_decode}
Whereas single-case substitution (\cref{fig:motivation}c) isolates one slot on an individual patient, \cref{tab:causal} tests across cohorts whether slots that linearly encode a report attribute also control the reader's answers for that attribute. For each of the 13 attributes, we fit standardised PCA and $\ell_2$-regularised logistic regression probes ($C=0.1$) inside patient-stratified five-fold cross-validation on mean patch features, compressed slots, projected tokens, and every second reader layer at visual and readout positions, scoring binary and one-vs-rest macro ROC-AUC. Fixing a linear probe family prevents confounding representation content with probe expressivity~\citep{Hewitt2019ControlProbes}, while single-slot counterfactual substitution (\cref{eq:patch}) complements amnesic feature removal~\citep{Elazar2021Amnesic} by inserting a real different-label donor vector on matching SlideBench-Evidence questions.

\paragraph{Tile coordinate verification and routed tissue.}
To confirm spatial alignment between stored CONCH feature rows and diagnostic WSI coordinates, re-encoding $20\times$ 224-px tiles across audited diagnostic slides retrieves the exact stored feature vector on every checked tile. For \cref{tab:slot_tissue_routing}, we pass all unique SlideBench-Evidence slides through the frozen \ours-8 top-2 router, normalise softmax weights within each slot as in forward inference, and assign patch tissue labels via the calibrated nine-class CONCH locator (\cref{sec:app_locator}; leading adjusted log-probability margin above 1.0, otherwise uncertain). Subtracting slide-wide class prevalence from each slot's routing-weighted share isolates tissue specialisation across slots.

\begin{table}[t]
\centering
\small
\setlength{\tabcolsep}{3pt}
\caption{\textbf{Visual attention and edge blocking on 48 questions.} Flip rates give the percentage of answers changed. Tumour and Context block the two slots with highest and lowest tumour mass.}
\label{tab:vision_text}
\resizebox{\linewidth}{!}{%
\begin{tabular}{lccccccccc}
\toprule
Model & $K$ & Q$\to$vis & Ref. & Above & Vis.\ flip & Prefix flip & Tumour $\Delta$ & Tumour flip & Context flip \\
\midrule
\rowcolor{PaperTableResult}
\ours & 8 & 14.3 & 12.0 & 45/48 & \TableBest{54.2} & 14.6 & \TableSecond{0.03} & 27.1 & 0.0\\
\rowcolor{white}
TC-SSA-8 & 8 & 12.3 & 12.0 & 25/48 & 39.6 & \TableBest{31.2} & \TableBest{0.14} & \TableSecond{29.2} & \TableSecond{12.5}\\
\rowcolor{PaperTableStripe}
\ours-SC & 8 & 8.3 & 12.0 & 0/48 & 43.8 & \TableSecond{16.7} & 0.01 & \TableBest{31.2} & \TableBest{14.6}\\
\rowcolor{white}
PathSelect & 32 & 9.6 & 35.1 & 0/48 & 33.3 & 12.5 & -0.01 & 6.2 & 0.0\\
\rowcolor{PaperTableStripe}
LoC-Path & 96 & 7.1 & 61.8 & 0/48 & 35.4 & 12.5 & 0.00 & 4.2 & 2.1\\
\rowcolor{white}
LoC-Path + \ours & 96 & 11.6 & 61.8 & 0/48 & \TableSecond{45.8} & 4.2 & 0.01 & 0.0 & 0.0\\
\rowcolor{PaperTableStripe}
Mean & 1 & 6.1 & 1.7 & 48/48 & 31.2 & 4.2 & -- & -- & --\\
\bottomrule
\end{tabular}}
\end{table}

\paragraph{Text-to-vision edge blocking and intervention statistics.}
\Cref{tab:vision_text} adapts prompt-wide attention and edge-knockout diagnostics~\citep{Zhang2024InfoFlow} to 48 Evidence questions, comparing question-to-slot attention against uniform token share and testing answer sensitivity when zeroing attention to all visual slots, an equal-length text prefix, or the two slots carrying the highest or lowest tumour mass. On \ours-8, blocking visual reads flips answers far more often than prefix blocking, whereas blocking the two most tumour-heavy slots produces only a minor log-probability drop; attribute-specific slot content drives answer support beyond raw tumour volume. Across the 780 single-slot substitutions (60 per attribute), the intervention effect in \cref{eq:patch} is evaluated in fp32 valid-option log-softmax space as
\begin{equation*}
\Delta_k = \log \frac{\pi_\theta(y \mid q, V_s)}{\pi_\theta(y \mid q, V_s^{\,k\leftarrow\bar s})}.
\end{equation*}
In \cref{tab:causal}, \ours-8 raises both slide-swap sensitivity and the within-attribute correlation between slot probe AUC and $|\Delta_k|$ while concentrating causal influence on fewer effective tokens than TC-SSA-8; permuting attribute labels lowers this correlation on both models, separating shared high-utility slots from attribute-specific routing. Attention and gradient-by-input correlations use the initial 150-question subset, and tumour--effect correlation covers the ten non-glioma tumour-associated attributes. Finally, a nonlinear gated-attention MIL probe~\citep{Ilse2018Attentionbased} on projected tokens provides an empirical reference across token budgets: whereas 32-token TC-SSA trails its token-probe reference in answer wAUC, eight-token TC-SSA-8 matches its own token-probe reference, while \ours-8 lifts both the token-probe reference and final answer wAUC above its probe point estimate (with a positive patient-paired interval over TC-SSA-8).

\subsection{Real-Slide Illustrations of Token Substitution}
\label{sec:visual_cases}
\begin{figure*}[t]
\centering
\includegraphics[width=\linewidth]{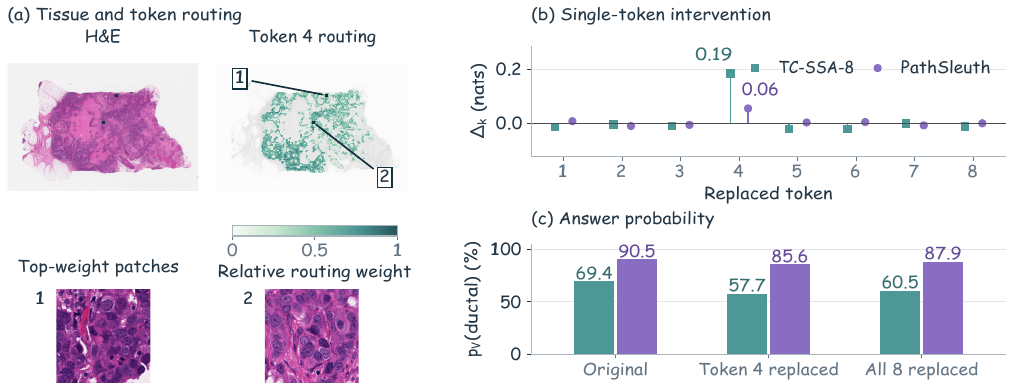}
\caption{\textbf{Token substitution on breast-subtype case 530.} Replacing token 4 lowers the ductal-answer probability from \caseFailureOriginal\% to \caseFailureSingle\%. Replacing all eight leaves it at \caseFailureAll\%. Patch selection follows \cref{fig:token_visual_evidence}.}
\label{fig:visual_failure}
\end{figure*}

\begin{figure*}[t]
\centering
\includegraphics[width=\linewidth]{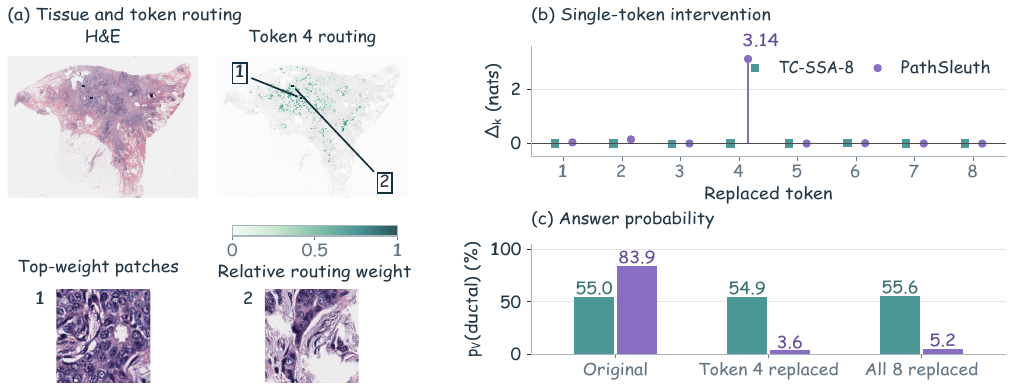}
\caption{\textbf{Tracing a ductal-subtype answer to token 4.} (a) Source slide, routing weights and two highest-weight patches. (b) Answer effects of single-token substitutions. (c) Replacing token 4 with its lobular-slide counterpart reduces ductal probability from \caseMainOriginal\% to \caseMainSingle\%.}
\label{fig:token_visual_evidence}
\end{figure*}

Cases 318 and 530 are the lowest-numbered locally indexed breast-subtype slides originally answered correctly by both eight-slot readers on which \ours-8 flips to the lobular donor (\cref{fig:token_visual_evidence}) or retains its ductal diagnosis (\cref{fig:visual_failure}). Substituting each donor slot individually and projecting within-slide normalised routing weights onto the SlideChat tile grid pairs spatial visualisation~\citep{Lu2021DataEfficient} with interventional verification~\citep{Adebayo2018Sanity}. In case 318, replacing slot~4 alone collapses ductal support in \ours-8 while leaving TC-SSA-8 nearly unchanged, showing targeted slot reliance. In case 530, ductal probability stays high in \ours-8 even after slot-4 and full-slide swaps; prompt priors outweigh slide cues.

\subsection{Patch-Gradient Attribution and Deletion}
\label{sec:patch_attribution}
\begin{figure*}[t]
\centering
\includegraphics[width=\linewidth]{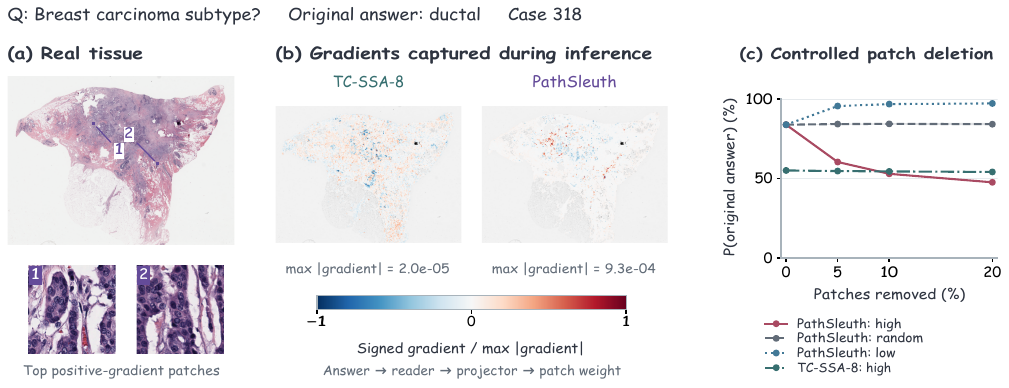}
\caption{\textbf{Patch gradients and deletion effects.} (a) Source slide and the two highest-gradient patches. (b) Signed inclusion gradients for the original predicted answer, scaled by each map's absolute maximum. (c) Answer support after deleting high- or low-gradient patches. Shading spans five random deletions of the same size.}
\label{fig:patch_attribution}
\end{figure*}

\begin{table*}[t]
\centering\small
\setlength{\tabcolsep}{5pt}
\caption{\textbf{Deleting 10\% of patches on 11 cases.} Drop in original-answer log probability (nats) for high or low signed gradients and five random masks (mean). $\dagger$ marks an incorrect original prediction.}
\label{tab:patch_cases}
\begin{tabularx}{\textwidth}{l *{9}{R}}
\toprule
& \multicolumn{3}{c}{TC-SSA-8} & \multicolumn{3}{c}{\ours} & \multicolumn{3}{c}{LoC-Path + \ours} \\
\cmidrule(lr){2-4}\cmidrule(lr){5-7}\cmidrule(lr){8-10}
QID & High & Rnd & Low & High & Rnd & Low & High & Rnd & Low \\
\midrule
\rowcolor{PaperTableStripe}
318 & \TableBest{0.011} & \TableSecond{0.002} & -0.004 & \TableBest{0.460} & \TableSecond{-0.006} & -0.144 & \TableBest{1.032} & \TableSecond{-0.011} & -0.110\\
\rowcolor{white}
325 & \TableBest{0.027}$^\dagger$ & \TableSecond{-0.006} & -0.024 & \TableBest{1.431} & \TableSecond{-0.002} & -0.153 & \TableBest{0.397} & \TableSecond{-0.004} & -0.067\\
\rowcolor{PaperTableStripe}
339 & \TableBest{0.022} & \TableSecond{0.003} & -0.094 & \TableBest{0.087} & \TableSecond{0.001} & -0.023 & \TableBest{0.675} & \TableSecond{-0.023} & -0.179\\
\rowcolor{white}
510 & \TableBest{0.023} & \TableSecond{0.001} & -0.021 & \TableBest{0.046} & \TableSecond{-0.004} & -0.110 & \TableBest{0.763}$^\dagger$ & \TableSecond{-0.002} & -0.346\\
\rowcolor{PaperTableStripe}
516 & \TableBest{0.013} & \TableSecond{-0.004} & -0.025 & \TableBest{0.065} & \TableSecond{0.001} & -0.024 & \TableBest{0.145} & \TableSecond{0.003} & -0.037\\
\rowcolor{white}
530 & \TableBest{0.106} & \TableSecond{0.002} & -0.005 & \TableBest{0.018} & \TableSecond{0.001} & -0.023 & \TableBest{1.287} & \TableSecond{-0.004} & -0.063\\
\rowcolor{PaperTableStripe}
883 & \TableBest{0.036} & \TableSecond{0.002} & -0.059 & \TableBest{0.494} & \TableSecond{-0.002} & -0.093 & \TableBest{0.244} & \TableSecond{-0.007} & -0.127\\
\rowcolor{white}
963 & \TableBest{0.017}$^\dagger$ & \TableSecond{-0.003} & -0.014 & \TableBest{1.777} & \TableSecond{-0.019} & -0.485 & \TableBest{1.605}$^\dagger$ & \TableSecond{0.030} & -0.413\\
\rowcolor{PaperTableStripe}
1012 & \TableBest{0.043} & \TableSecond{-0.007} & -0.021 & \TableBest{0.952}$^\dagger$ & \TableSecond{-0.012} & -0.279 & \TableBest{0.372} & \TableSecond{0.044} & -0.066\\
\rowcolor{white}
1110 & \TableBest{0.114} & \TableSecond{0.001} & -0.004 & \TableBest{0.037} & \TableSecond{0.002} & -0.024 & \TableBest{2.964}$^\dagger$ & \TableSecond{-0.065} & -0.429\\
\rowcolor{PaperTableStripe}
1128 & \TableBest{0.155} & \TableSecond{-0.004} & -0.100 & \TableBest{0.276} & \TableSecond{-0.001} & -0.090 & \TableBest{1.050} & \TableSecond{-0.006} & -0.078\\
\midrule
\rowcolor{white}
Mean & \TableBest{0.052} & \TableSecond{-0.001} & -0.034 & \TableBest{0.513} & \TableSecond{-0.004} & -0.132 & \TableBest{0.958} & \TableSecond{-0.004} & -0.174\\
\bottomrule
\end{tabularx}
\end{table*}

\begin{figure*}[t]
\centering
\includegraphics[width=0.92\linewidth]{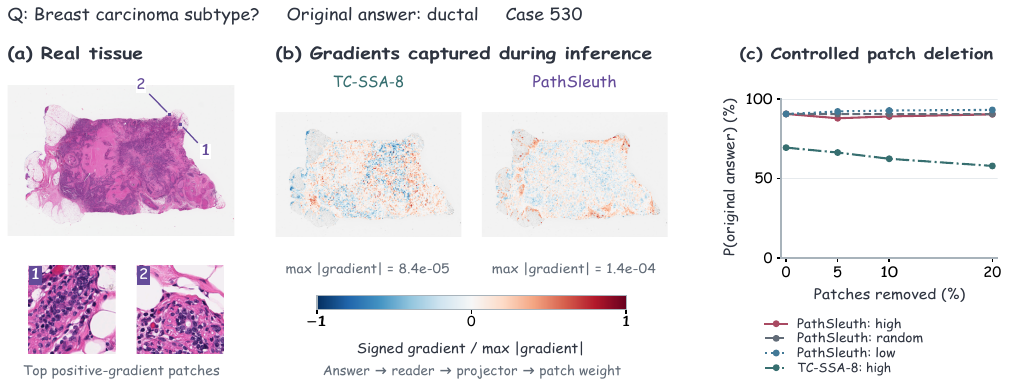}
\caption{\textbf{Patch gradients and deletion effects on case 530.} Each deletion uses the original predicted answer as its target. Gradient colours are scaled by each map's absolute maximum, as in \cref{fig:patch_attribution}.}
\label{fig:patch_attribution_failure}
\end{figure*}

Across all 11 locally geometry-indexed breast-subtype slides (retaining original option orders and initial predictions), we multiply patch $i$'s routing weight (or the four-patch merger input in 96-token LoC-Path) by an inclusion scalar $a_i$ evaluated at $a=\mathbf{1}$ and compute the predicted-option log-probability gradient across the full visual-to-reader path:
\begin{equation*}
g_i = \left.\frac{\partial \log p_V(\hat y \mid q, a)}{\partial a_i}\right|_{a=\mathbf{1}}.
\end{equation*}
Nominal probabilities match standard inference on every run. Following region-deletion saliency protocols~\citep{Petsiuk2018RISE,Adebayo2018Sanity}, we zero validity masks for the top 5\%, 10\%, or 20\% positively ranked patches, bottom negatively ranked patches, and five seeded random masks of matched size per slide without rerunning CONCH (\cref{fig:patch_attribution,fig:patch_attribution_failure}). In \cref{tab:patch_cases}, removing high-gradient patches produces a stronger support drop than random deletion on every slide for every model, and \ours-8 shows an order-of-magnitude larger mean drop than TC-SSA-8, separating focal visual reliance in case 318 from diffuse insensitivity in case 530.

\subsection{Recorded Attention and Layer Blocking}
\label{sec:attention_capture}
\begin{table*}[t]
\centering\small
\setlength{\tabcolsep}{5pt}
\caption{\textbf{Visual attention on 11 cases.} Q is question-to-slot attention (\%). V and P give original-answer log-probability drops (nats) after visual and prefix blocks. $\dagger$ marks an incorrect prediction.}
\label{tab:attention_cases}
\begin{tabularx}{\textwidth}{l *{8}{R}}
\toprule
& \multicolumn{3}{c}{TC-SSA-8} & \multicolumn{3}{c}{\ours} & \multicolumn{2}{c}{LoC-Path + \ours} \\
\cmidrule(lr){2-4}\cmidrule(lr){5-7}\cmidrule(lr){8-9}
QID & Q & V & P & Q & V & P & Q & V \\
\midrule
\rowcolor{PaperTableStripe}
318 & 11.5 & \TableSecond{-0.052} & \TableBest{0.129} & 11.6 & \TableBest{0.344} & \TableSecond{0.084} & 7.9 & 0.202\\
\rowcolor{white}
325 & 10.9$^\dagger$ & \TableSecond{-0.278} & \TableBest{-0.207} & 12.0 & \TableBest{0.918} & \TableSecond{-0.229} & 10.9 & 1.270\\
\rowcolor{PaperTableStripe}
339 & 12.1 & \TableSecond{-0.029} & \TableBest{0.159} & 11.5 & \TableBest{0.441} & \TableSecond{0.117} & 13.6 & 0.166\\
\rowcolor{white}
510 & 11.7 & \TableSecond{-0.033} & \TableBest{0.212} & 9.9 & \TableBest{0.475} & \TableSecond{0.108} & 8.9$^\dagger$ & 0.819\\
\rowcolor{PaperTableStripe}
516 & 11.7 & \TableBest{-0.020} & \TableSecond{-0.204} & 11.9 & \TableBest{0.562} & \TableSecond{0.064} & 10.5 & 0.327\\
\rowcolor{white}
530 & 11.5 & \TableBest{0.146} & \TableSecond{-0.214} & 11.7 & \TableBest{0.444} & \TableSecond{0.032} & 10.9 & 0.326\\
\rowcolor{PaperTableStripe}
883 & 11.2 & \TableBest{0.003} & \TableSecond{0.002} & 12.2 & \TableBest{0.381} & \TableSecond{0.205} & 9.2 & 0.205\\
\rowcolor{white}
963 & 11.8$^\dagger$ & \TableSecond{-0.039} & \TableBest{0.149} & 13.1 & \TableBest{0.601} & \TableSecond{-0.097} & 8.2$^\dagger$ & -0.299\\
\rowcolor{PaperTableStripe}
1012 & 11.5 & \TableBest{0.121} & \TableSecond{0.048} & 12.4 & \TableSecond{-0.301} & \TableBest{0.228} & 8.3 & 0.251\\
\rowcolor{white}
1110 & 11.6 & \TableBest{0.072} & \TableSecond{0.046} & 11.8 & \TableBest{0.463} & \TableSecond{0.043} & 8.0$^\dagger$ & 1.792\\
\rowcolor{PaperTableStripe}
1128 & 11.7 & \TableBest{0.046} & \TableSecond{-0.289} & 11.8 & \TableBest{0.402} & \TableSecond{0.048} & 8.7 & 0.315\\
\midrule
\rowcolor{white}
Mean & 11.6 & \TableBest{-0.006} & \TableSecond{-0.016} & 11.8 & \TableBest{0.430} & \TableSecond{0.055} & 9.6 & 0.489\\
\bottomrule
\end{tabularx}
\end{table*}

\begin{figure*}[t]
\centering
\includegraphics[width=\linewidth]{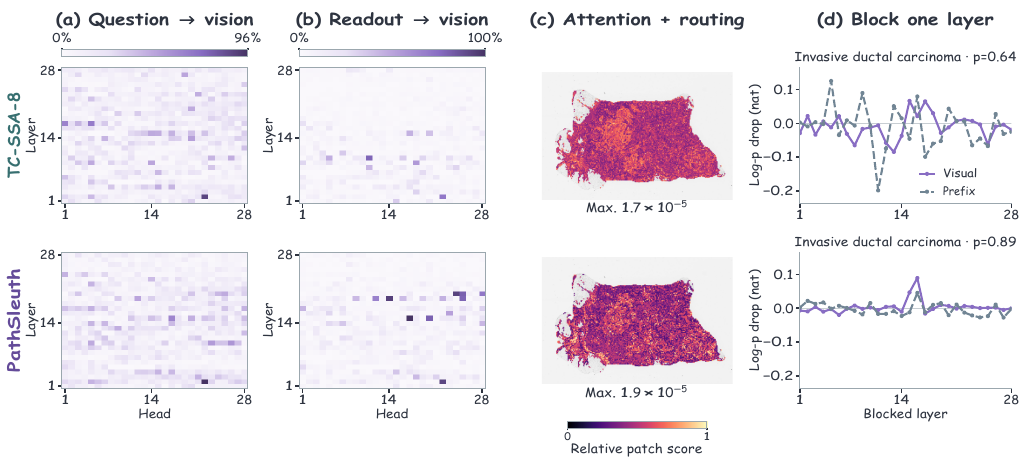}
\caption{\textbf{Attention and layer blocking on case 530.} Head-level reads, routing maps and original-answer effects use the protocol of \cref{fig:attention_capture}.}
\label{fig:attention_failure}
\end{figure*}
\begin{figure*}[t]
\centering
\includegraphics[width=.83\linewidth]{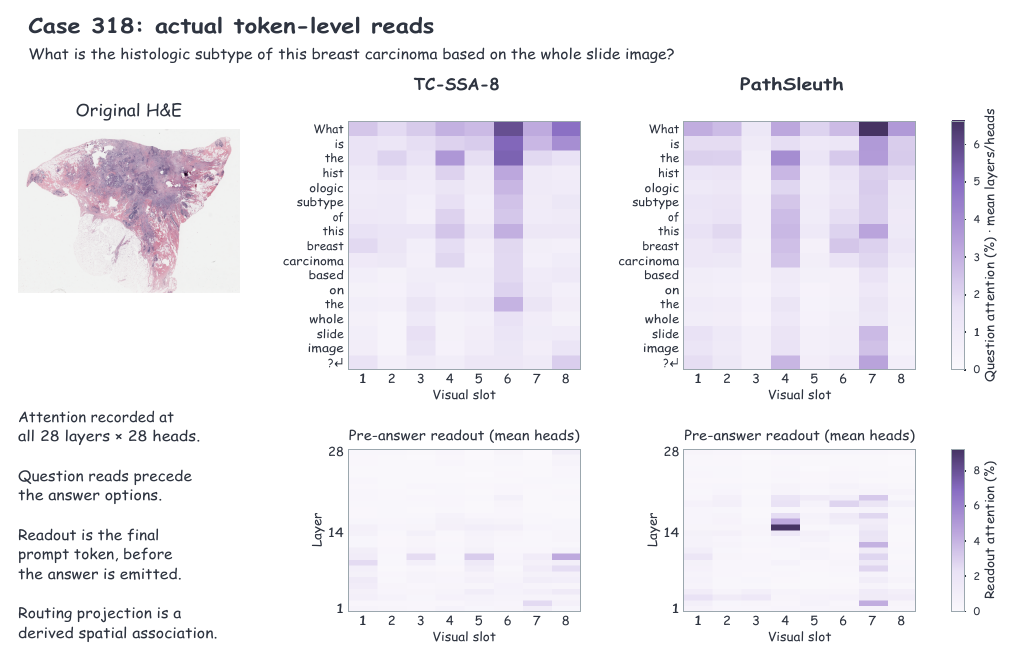}
\caption{\textbf{Question and pre-answer attention on case 318.} Top panels average each question token's attention to each slot over heads and layers. Bottom panels show the final prompt token's attention by layer, averaged over heads.}
\label{fig:attention_tokens}
\end{figure*}
\begin{figure*}[t]
\centering
\includegraphics[width=\linewidth]{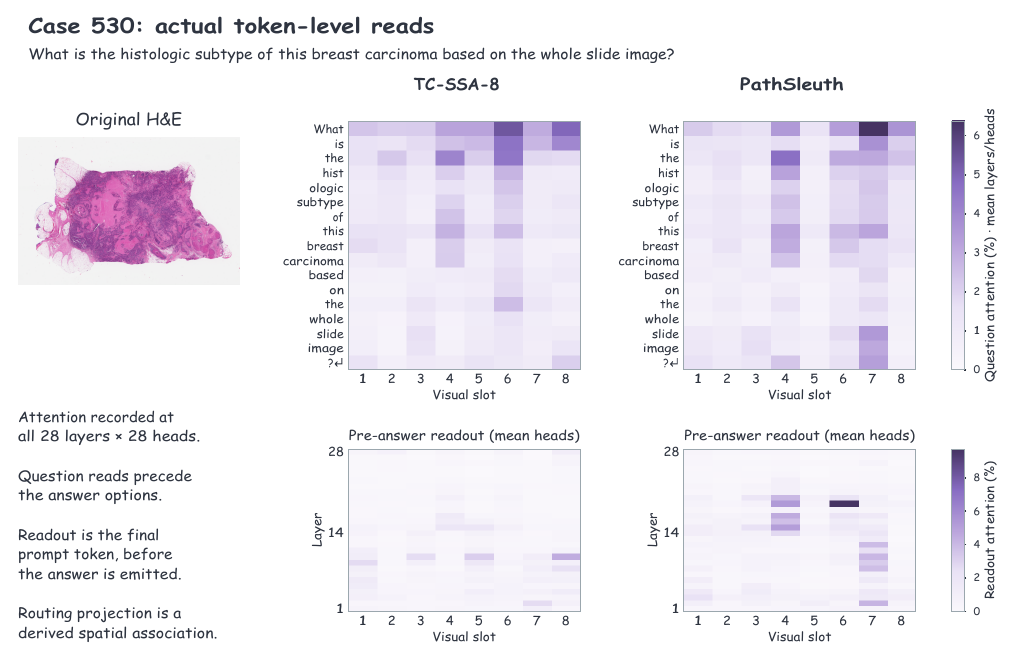}
\caption{\textbf{Token-level attention on case 530.} Panels use the measurements and averaging conventions of \cref{fig:attention_tokens}.}
\label{fig:attention_tokens_failure}
\end{figure*}

On the same 11 breast-subtype slides, eager-attention forward passes record full layer-by-head attention tensors, pre-RoPE query and key projections, value vectors, visual slots, and routing weights (\cref{tab:attention_cases,fig:attention_failure,fig:attention_tokens,fig:attention_tokens_failure}). In the tokenised prompt, visual slots occupy a contiguous prefix before question tokens, option text, and the final pre-answer readout position; eager and FlashAttention predictions agree across nearly all model-case runs. Suppressing causal query-to-key edges via pre-softmax hooks (verified to zero masked attention exactly and recover nominal probabilities when removed) pairs raw attention maps with causal edge tests~\citep{Jain2019Attention,Serrano2019Attention,Abnar2020AttentionFlow}. Although TC-SSA-8 and \ours-8 allocate comparable raw attention to the visual slots, blocking all text-to-visual edges causes a sharp answer drop only in \ours-8, far above an equal-length prefix control across almost every slide. Blocking only question queries or only the final readout token has a much smaller effect because visual features propagate through intermediate option tokens, while blocking the two highest-attended slots outweighs the bottom two on all 11 slides.

\begin{table}[t]
\centering\small
\setlength{\tabcolsep}{2.2pt}
\caption{\textbf{Blocking visual reads on 11 breast-subtype cases.} Entries are mean drops in benchmark-answer log probability (nats). Each layer group is blocked jointly. Intervals appear in \cref{tab:attention_inference}.}
\label{tab:attention_focus}
\begin{tabularx}{\linewidth}{Lccc}
\toprule
Blocked reads & TC-SSA-8 & \ours-8 & \shortstack{\rule{0pt}{2.1ex}LoC-Path + \\ \ours} \\
\midrule
\rowcolor{PaperTableStripe}
All text positions & 0.047 & \TableBest{0.430} & \TableSecond{0.169}\\
\rowcolor{white}
Question positions only & \TableSecond{-0.017} & \TableBest{0.009} & -0.033\\
\rowcolor{PaperTableStripe}
Answer readout only & \TableSecond{-0.003} & \TableBest{0.041} & -0.019\\
\rowcolor{white}
Visual minus prefix & \TableSecond{0.051} & \TableBest{0.375} & --\\
\midrule
\rowcolor{PaperTableStripe}
Layers 1--7 & \TableSecond{0.003} & \TableBest{0.009} & -0.003\\
\rowcolor{white}
Layers 8--14 & -0.044 & \TableBest{0.061} & \TableSecond{0.042}\\
\rowcolor{PaperTableStripe}
Layers 15--21 & 0.080 & \TableBest{0.304} & \TableSecond{0.146}\\
\rowcolor{white}
Layers 22--28 & -0.015 & \TableBest{0.003} & \TableSecond{-0.005}\\
\bottomrule
\end{tabularx}
\end{table}

\begin{table*}[t]
\centering\small
\setlength{\tabcolsep}{6pt}
\caption{\textbf{Visual dependence and its layer distribution.} Mean log-probability contrasts (nats) with exploratory pointwise paired-patient percentile 95\% intervals from 100,000 resamples of the 11 fixed cases. Native target is each model\textquotesingle s original predicted answer; GT fixes the same benchmark answer across models. Layer contrasts concern \ours.}
\label{tab:attention_inference}
\begin{tabularx}{\textwidth}{>{\hsize=1.24\hsize}L >{\hsize=0.88\hsize}Y >{\hsize=0.88\hsize}Y}
\toprule
Contrast & Native target & Common GT \\
\midrule
\rowcolor{PaperTableStripe}
\ours: visual block & 0.430 [0.251, 0.578] & \TableBest{0.430} [0.251, 0.578]\\
\rowcolor{white}
TC-SSA-8: visual block & -0.006 [-0.074, 0.052] & 0.047 [-0.000, 0.102]\\
\rowcolor{PaperTableStripe}
LoC-Path + \ours: visual block & 0.489 [0.189, 0.846] & \TableSecond{0.169} [-0.130, 0.461]\\
\rowcolor{white}
\ours{} minus TC-SSA-8 & 0.436 [0.215, 0.646] & 0.383 [0.197, 0.515]\\
\rowcolor{PaperTableStripe}
\ours: visual minus prefix & 0.375 [0.141, 0.595] & 0.375 [0.141, 0.595]\\
\rowcolor{white}
Difference after prefix control & 0.365 [0.091, 0.631] & 0.325 [0.069, 0.573]\\
\rowcolor{PaperTableStripe}
Layers 15--21 minus 1--7 & 0.295 [0.103, 0.511] & 0.295 [0.103, 0.511]\\
\rowcolor{white}
Layers 15--21 minus 8--14 & 0.243 [0.043, 0.458] & 0.243 [0.043, 0.458]\\
\rowcolor{PaperTableStripe}
Layers 15--21 minus 22--28 & 0.301 [0.102, 0.525] & 0.301 [0.102, 0.525]\\
\bottomrule
\end{tabularx}
\end{table*}

\paragraph{Bootstrap contrasts across layers.}
Jointly resampling the 11 patients 100,000 times (\cref{tab:attention_focus,tab:attention_inference}) confirms positive visual dependence in \ours-8 over both TC-SSA-8 and the prefix control, whether evaluated on each model's native prediction or on shared ground-truth targets. Partitioning the 28 reader layers into four seven-layer blocks isolates visual readout sharply in middle-to-late layers 15--21, whose knockout effect is larger than every earlier and later block with paired intervals excluding zero.

\subsection{Slot Specificity and Readout Depth}
\label{sec:vet_extended}
\begin{figure*}[t]
\centering
\includegraphics[width=\linewidth,trim=0 232 0 0,clip]{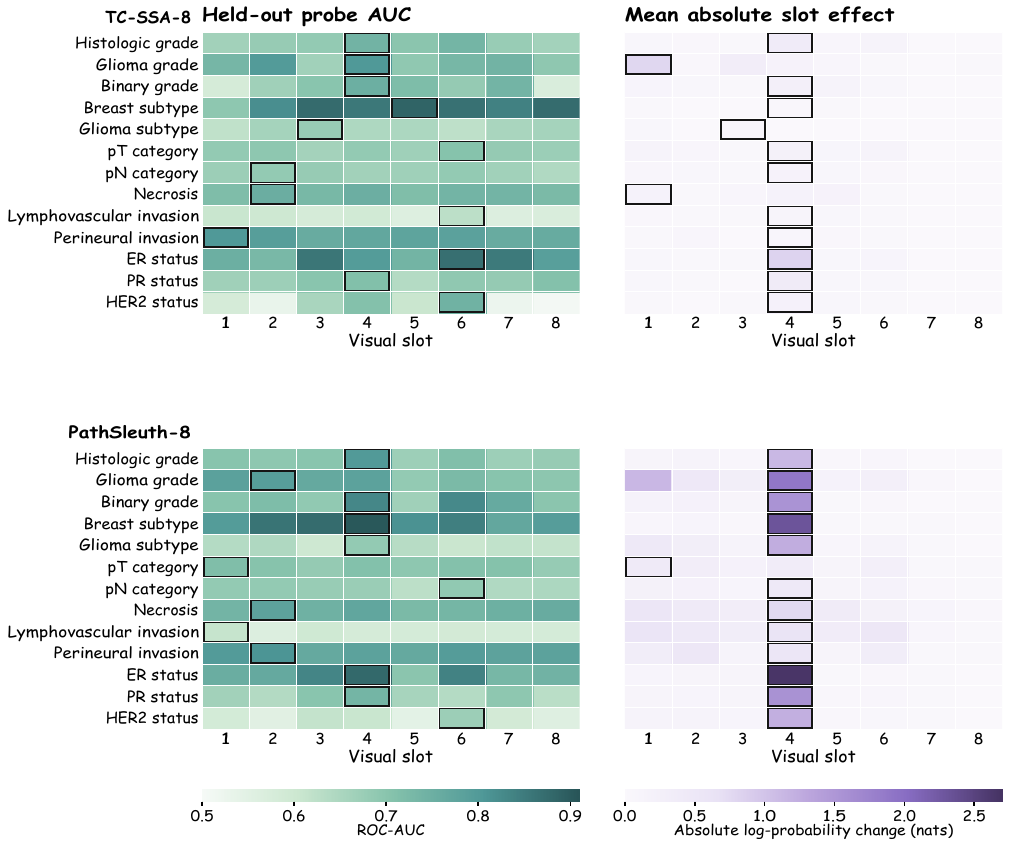}\\[-2pt]
\includegraphics[width=\linewidth,trim=0 0 0 205,clip]{figures/print/fig_vet_specificity.pdf}
\caption{\textbf{Slot-level encoding and causal answer effect across 13 report attributes.} Left: five-fold held-out linear probe ROC-AUC per slot. Right: mean absolute log-probability shift $|\Delta_k|$ (nats) under single-slot substitution across 60 questions per attribute (721/780 same-cancer donors). Bold boxes mark row maxima on shared cross-model scales.}
\label{fig:vet_specificity}
\end{figure*}
\begin{figure*}[t]
\centering
\includegraphics[width=\linewidth]{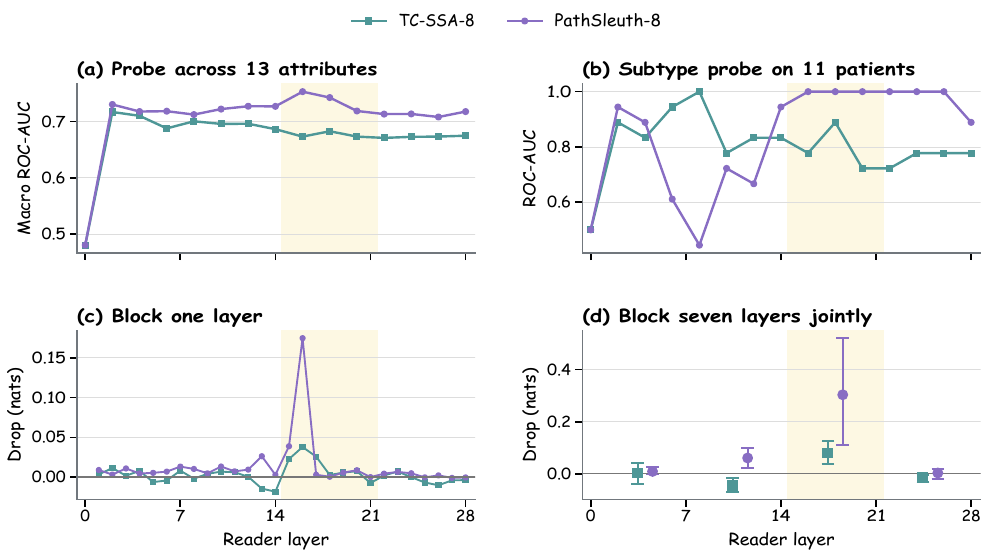}
\caption{\textbf{Attribute decodability versus layerwise visual readout.} (a) Five-fold macro probe AUC across 13 attributes at the pre-answer token every two layers. (b) Held-out breast-subtype probe AUC (trained on 157 external patients, scored on the 11 captured slides). (c,d) Benchmark-answer log-probability drop under single-layer or seven-layer visual blocking (95\% bootstrap CIs; shading marks layers 15--21).}
\label{fig:vet_depth}
\end{figure*}
\Cref{fig:vet_specificity} compares held-out slot probe AUC against single-slot $|\Delta_k|$ across all 13 attributes. In \ours-8, encoding and causal impact align closely by biological structure: tumour-enriched slot~4 anchors histologic grade and subtypes, slot~1 leads both decodability and answer impact for pT category, and slot~6 and slot~4 split peak decodability and causal influence on pN category. Across reader depth (\cref{fig:vet_depth}), attribute information becomes linearly accessible at the pre-answer token in the earliest layers, whereas single-layer visual blocking peaks later in the middle-to-late stack around layer~16 with minimal prefix-control shift. Middle-to-late layers therefore integrate visual evidence into answer formation well after visual attributes first appear linearly in the residual stream. In cumulative $m$-slot exchanges on the 780 tracing pairs (\cref{fig:vet_cumulative}), slots are ranked by nested inner-fold probe AUC excluding both source and donor outer folds, and cumulative restoration from a full donor swap $D$ back toward the original slide $O$ is measured by
\begin{equation*}
\rho_{\mathrm{restore}}(m) = \frac{\sum_i \bigl[L_i(R_m) - L_i(D)\bigr]}{\sum_i \bigl[L_i(O) - L_i(D)\bigr]}.
\end{equation*}
Deleting the top three probe-ranked slots removes far more answer support than random or bottom-ranked deletions, and restoring the top three slots recovers 81.0\% of the full-swap loss in \ours-8, clearly ahead of frozen TC-SSA-8 across patient resamples.

\subsection{Probe Support on Incorrect Answers}
\label{sec:vet_errors}
\begin{table}[t]\centering\footnotesize
\setlength{\tabcolsep}{3pt}
\caption{\textbf{Probe support on incorrect answers.} Counts among each model's errors on all 3,994 original questions. Small denotes an absolute full-swap change of at most 0.10 nat on the true option.}
\label{tab:vet_errors}
\begin{tabularx}{\linewidth}{Lrrr}\toprule
 & TC-SSA-8 & \ours-8 & \ours-SC \\ \midrule
\rowcolor{PaperTableStripe}
All probes low & 340 & 239 & 286\\
\rowcolor{white}
High, small swap & 73 & 18 & 24\\
\rowcolor{PaperTableStripe}
High, larger swap & 188 & 285 & 256\\
\rowcolor{white}
Intermediate & 1,000 & 864 & 886\\
\rowcolor{PaperTableStripe}
Unsupported label & 2 & 2 & 2\\
\midrule
\rowcolor{white}
All errors & 1,603 & \TableBest{1,408} & \TableSecond{1,454}\\
\bottomrule\end{tabularx}
\end{table}

To separate visual compression loss from reader under-use among all incorrect original answers on SlideBench-Evidence, we pair each error with a different-label donor and evaluate outer-fold-excluded probes over the $J$ supported labels at uniform chance $u = 1/J$ (\cref{tab:vet_errors}). Errors are partitioned into \emph{all probes low} (every slot assigns true-label probability at or below $u$) and \emph{high support} (the top-ranked slot reaches at least halfway from $u$ to certainty), subdivided by whether a full donor swap shifts true-option log probability by at most 0.10 nat (\emph{small swap}). In \cref{tab:vet_errors}, \ours-8 cuts high-support, small-swap errors (cases where decodable slot evidence fails to move the reader) more than fourfold relative to TC-SSA-8, converting most of those baseline failures into correct predictions; the same reduction holds across support and swap thresholds and on the balanced subset.

\section{Supervision Signals and Domain Reader}
\label{sec:app_design}
\newcommand{\vetArmAAuc}{0.763}
\newcommand{\vetArmARho}{0.498}
\newcommand{\vetArmBAuc}{0.762}
\newcommand{\vetArmBRho}{0.526}
\newcommand{\vetArmCAuc}{0.766}
\newcommand{\vetArmCRho}{0.510}
\newcommand{\vetArmDAuc}{0.765}
\newcommand{\vetArmDRho}{0.454}

\begin{table*}[t]
\centering
\small
\setlength{\tabcolsep}{7pt}
\caption{\textbf{Matched one-epoch supervision arms.} Validation covers 2,081 real questions, 1,599 same-question pairs, 645 regional pairs and 4,218 knowledge questions (\%). Best-slot AUC averages 13 attributes. Alignment ($\rho$) uses 780 test questions.}
\label{tab:g2_matched}
\begin{tabularx}{\textwidth}{Lrrrrrr}
\toprule
Recipe & Real & Paired correct & Donor following & Knowledge & Best-slot AUC & $\rho$ \\
\midrule
\rowcolor{white}
A: base & \TableBest{65.64} & 25.89 & 10.54 & \TableBest{88.35} & \vetArmAAuc & \vetArmARho\\
\rowcolor{PaperTableStripe}
B: A + paired CE & 63.00 & \TableBest{34.33} & 12.87 & 88.13 & \vetArmBAuc & \TableBest{\vetArmBRho}\\
\rowcolor{white}
C: B + relation views & \TableSecond{64.10} & \TableSecond{31.21} & \TableSecond{21.09} & \TableSecond{88.19} & \TableBest{\vetArmCAuc} & \TableSecond{\vetArmCRho}\\
\rowcolor{PaperTableResult}
D: C + preserve KL & 63.38 & 28.96 & \TableBest{22.48} & 88.18 & \TableSecond{\vetArmDAuc} & \vetArmDRho\\
\bottomrule
\end{tabularx}
\end{table*}

\begin{table*}[t]
\centering
\small
\setlength{\tabcolsep}{3pt}
\caption{\textbf{Supervision recipes at $K{=}8$ on $\mathcal{D}_\text{ev+}$.} E2E: trainable visual modules; Lab: attribute heads; Anc: preserve/alter; Evd: keep/transplant. The first five rows compare one-epoch supervision variants; the final row reports the full two-epoch \ours-8 schedule (\cref{sec:app_design}). Tr.: unconditional donor following on common pairs.}
\label{tab:ablation}
\begin{tabularx}{\textwidth}{*{11}{Y}}
\toprule
E2E & Lab & Anc & Evd & SB & VR & VR Circ. & $\Delta_\text{swap}$ & Pair & wAUC & Tr. \\
\midrule
\rowcolor{white}
 &  &  &  & 67.0 & 41.6 & 18.7 & 4.2 & 19.7 & .649 & 8.9\\
\rowcolor{PaperTableStripe}
\cmark &  &  &  & 67.4 & 42.5 & 20.4 & 5.3 & 19.6 & .656 & 9.5\\
\rowcolor{white}
\cmark & \cmark &  &  & 69.3 & 47.3 & 25.3 & 8.4 & 24.7 & \TableBest{.702} & 10.7\\
\rowcolor{PaperTableStripe}
\cmark &  & \cmark &  & 67.5 & 44.1 & 17.9 & 7.0 & 28.7 & .658 & 10.1\\
\rowcolor{white}
\cmark & \cmark & \cmark &  & \TableBest{69.8} & \TableSecond{48.6} & \TableSecond{26.8} & \TableBest{10.2} & \TableSecond{32.1} & .692 & \TableSecond{15.6}\\
\rowcolor{PaperTableResult}
\cmark & \cmark & \cmark & \cmark & \TableSecond{69.6} & \TableBest{49.6} & \TableBest{31.0} & \TableSecond{10.0} & \TableBest{33.3} & \TableSecond{.696} & \TableBest{32.0}\\
\bottomrule
\end{tabularx}
\end{table*}

This section isolates the supervision objectives and reader-initialisation choices behind \ours. We first analyse a strictly update-matched four-arm study (\cref{tab:g2_matched}) together with the full-recipe ablation progression and paired cross-entropy bound (\cref{tab:ablation}), then examine the pathology-pretrained reader variant \ours-SC and its component ablations in \cref{sec:official_variant}.

\paragraph{Matched four-arm supervision study.}
To isolate each objective term without schedule or compute confounds, the four arms in \cref{tab:g2_matched} share Stage-1 weights, attribute heads, Qwen2.5-7B initialisation, seed 43, and 968 Stage-2 optimizer updates across two GPUs (micro-batch 2, accumulation 32), running identical forward views per step while zeroing inactive loss weights. Evaluation spans held-out Stage-2 validation patients (real questions, same-question pairs, regional pairs, and knowledge items) alongside the disjoint 780-question test tracing cohort. Because every arm retains $\mathcal{L}_{\mathrm{lab}}$, best-slot linear probe AUC stays tightly matched across all four arms, so the observed differences come from how the reader uses visual tokens with encoder capacity held fixed. Adding Alter cross-entropy (Arm~B versus Arm~A) raises paired correctness and slot--answer alignment, whereas adding Preserve, Keep, and Transplant views (Arms~C and~D) roughly doubles joint regional donor following while keeping knowledge retention stable.

\paragraph{Full-recipe progression and paired loss bound.}
\label{sec:paired_bound}
\Cref{tab:ablation} tracks one-epoch component additions on $\mathcal{D}_{\mathrm{ev+}}$ alongside the full two-epoch \ours-8 schedule. Unfreezing visual slots alone leaves paired correctness near the frozen baseline; standard autoregressive instruction tuning does not compel the reader to use compressed visual tokens. Combining attribute heads with Preserve and Alter delivers most of the gain in paired correctness within a single epoch, and adding Keep and Transplant under the full schedule produces the main jump in regional donor following, matching a dedicated transplant-only control on tumour versus non-tumour specificity. The paired objective also imposes a direct lower bound on slide-agnostic guessing. On any Alter pair sharing question and option order with distinct target labels, a slide-independent distribution assigning probabilities $u$ and $v$ to the two targets satisfies
\begin{equation*}
u + v \le 1 \implies uv \le \frac{1}{4}, \qquad -\log(uv) \ge 2\log 2,
\end{equation*}
attained at $u=v=1/2$. Lowering summed paired cross-entropy below $\log 2$ per item therefore requires conditioning on slide-specific visual evidence.

\subsection{Released-Reader Variant (\ours-SC)}
\label{sec:official_variant}
\newcommand{\componentPermCircGain}{19.44}
\newcommand{\componentPermCircLo}{16.90}
\newcommand{\componentPermCircHi}{22.11}
\newcommand{\componentReplayVRGain}{0.83}
\newcommand{\componentReplayVRLo}{-0.88}
\newcommand{\componentReplayVRHi}{2.36}
\newcommand{\componentPermCirc}{41.3}
\newcommand{\componentPermEvidence}{63.8}
\newcommand{\componentPermWauc}{0.673}
\newcommand{\componentPermPair}{31.7}
\newcommand{\componentPermTrans}{30.8}
\newcommand{\componentPermMic}{50.8}
\newcommand{\componentPermMicSwap}{50.9}
\newcommand{\componentReplayCirc}{22.6}
\newcommand{\componentReplayEvidence}{64.5}
\newcommand{\componentReplayWauc}{0.691}
\newcommand{\componentReplayPair}{31.7}
\newcommand{\componentReplayTrans}{31.5}
\newcommand{\componentReplayMic}{50.7}
\newcommand{\componentReplayMicSwap}{50.4}
\newcommand{\componentBothCirc}{42.0}
\newcommand{\componentBothEvidence}{63.6}
\newcommand{\componentBothWauc}{0.669}
\newcommand{\componentBothPair}{28.9}
\newcommand{\componentBothTrans}{26.4}
\newcommand{\componentBothMic}{51.4}
\newcommand{\componentBothMicSwap}{51.7}
\begin{table*}[t]
\centering\small
\caption{\textbf{Permutation and replay} ($K{=}8$). The last three rows share the two-GPU setup; replay adds 25k examples and updates. $^\dagger$Historical one-GPU reference. Same-question Pair fixes the question and option order on 3,811 pairs; unconditional transplant uses 1,349 eligible pairs.}
\label{tab:components}
\resizebox{\textwidth}{!}{%
\begin{tabular}{lccrrrrrrr}
\toprule
Variant & Perm. & Replay & SB 4-ord. & VR 4-ord. & VR Circ. & Evidence & wAUC & Same-question Pair & Transplant \\
\midrule
\rowcolor{PaperTableStripe}
Earlier reference$^\dagger$ & No & No & 68.5 & 48.3 & 19.5 & 62.7 & 0.671 & \TableSecond{30.4} & \TableBest{32.4}\\
\rowcolor{white}
Permutation only & Yes & No & \TableSecond{73.5} & \TableSecond{53.2} & \TableSecond{41.3} & \TableSecond{63.8} & \TableSecond{0.673} & \TableBest{31.7} & 30.8\\
\rowcolor{PaperTableStripe}
Replay only & No & Yes & 69.6 & 49.5 & 22.6 & \TableBest{64.5} & \TableBest{0.691} & \TableBest{31.7} & \TableSecond{31.5}\\
\rowcolor{PaperTableResult}
\ours-SC & Yes & Yes & \TableBest{74.2} & \TableBest{54.0} & \TableBest{42.0} & 63.6 & 0.669 & 28.9 & 26.4\\
\bottomrule
\end{tabular}}
\end{table*}

\ours-SC initialises the reader from SlideChat's pathology-tuned checkpoint~\citep{Chen2025SlideChat}, aligns an eight-slot TC-SSA and projector interface, and trains for one instruction epoch at batch size 128 (two workers, micro-batch 2, accumulation 32) with option-permutation supervision (content-aligned KL and permuted option cross-entropy) and 25,000 open-ended SlideInstruct replay examples. Removing each component under identical two-GPU settings (\cref{tab:components}) clarifies their distinct roles: option-permutation supervision drives order-consistent CircularEval, open-ended replay maximises single-slide Evidence accuracy and within-cancer discrimination, and combining both achieves the highest SlideBench-VR accuracy and CircularEval while moderating contrastive pair and regional donor sensitivity.

\section{Regional Controls and Dataset Transfer}
\label{sec:app_locator}
\label{sec:app_robustness}
\begin{table*}[t]\centering\small
\setlength{\tabcolsep}{10pt}
\caption{\textbf{Feature-exchange controls for the main models.} Both views must follow the donor (tumour or non-tumour exchange). Same corr./unch. measure joint correctness or unchanged predictions after same-label exchange. $\Delta$logp compares tumour and non-tumour donors.}
\label{tab:transplant_controls}
\begin{tabularx}{\textwidth}{Lrrrrr}
\toprule
Model & Tumour & Same corr. & Same unch. & Non-tumour & $\Delta$logp \\ \midrule
\rowcolor{PaperTableResult}
\ours & \TableBest{31.4} & \TableSecond{56.6} & \TableSecond{56.4} & 6.6 & \TableSecond{0.47}\\
\rowcolor{white}
\ours-SC & 25.8 & 49.8 & 43.3 & \TableSecond{4.5} & \TableBest{2.85}\\
\rowcolor{PaperTableStripe}
LoC-Path + \ours & \TableSecond{28.7} & 54.0 & 56.1 & 9.0 & 0.30\\
\rowcolor{white}
TC-SSA-8 & 8.7 & \TableBest{64.7} & \TableBest{78.2} & \TableBest{3.7} & 0.08\\
\bottomrule
\end{tabularx}
\end{table*}

\begin{table*}[!t]
\begin{minipage}[t]{\columnwidth}

\centering\small
\setlength{\tabcolsep}{2.5pt}
\caption{\textbf{Responses to visual interventions.} Keep measures same-label accuracy. Don./Ctx. require both views to follow the donor. All rows share coverage within each intervention.}
\label{tab:evidence_detailed}
\resizebox{\linewidth}{!}{\begin{tabular}{l cc cc cc}
\toprule
& & & \multicolumn{2}{c}{Transplant} & \multicolumn{2}{c}{Del.} \\
\cmidrule(lr){4-5}\cmidrule(lr){6-7}
Method & Keep & Sel. & Don. & Ctx. & tum. & ctx. \\
\midrule
\rowcolor{PaperTableStripe}
SlideChat$^\dagger$ & 40.0 & 1.5 & 1.7 & 3.4 & 4.2 & \TableSecond{1.2}\\
\rowcolor{white}
SLICEChat$^\dagger$ & 52.3 & 6.4 & 6.9 & 7.9 & 14.6 & 8.5\\
\rowcolor{PaperTableStripe}
HistoSelect$^\dagger$ & 46.7 & 0.4 & 2.1 & \TableSecond{1.1} & 1.8 & 1.8\\
\rowcolor{white}
Mean pooling & 53.0 & 1.6 & 4.9 & 8.5 & 5.8 & 2.8\\
\rowcolor{PaperTableStripe}
TCP-LLaVA$^\ast$ & 49.2 & -0.3 & 0.0 & \TableBest{0.0} & 0.7 & \TableBest{0.5}\\
\rowcolor{white}
LoC-Path$^\ast$ & 54.2 & 7.2 & 7.2 & 14.5 & 12.4 & 4.8\\
\rowcolor{PaperTableStripe}
LoC-Path + \ours & 61.5 & \TableSecond{10.6} & \TableSecond{29.4} & 9.1 & \TableSecond{22.1} & 5.7\\
\rowcolor{white}
PathSelect$^\ast$ & 57.9 & 5.5 & 12.3 & 6.4 & 17.6 & 2.0\\
\rowcolor{PaperTableStripe}
TC-SSA-32 & 58.5 & 3.5 & 8.0 & 11.9 & 9.2 & 4.3\\
\rowcolor{white}
TC-SSA-8 & 58.9 & 5.7 & 8.9 & 9.9 & 12.0 & 3.9\\
\rowcolor{PaperTableStripe}
\ours, slide relations & 60.2 & 9.8 & 15.6 & 7.7 & 19.5 & 5.4\\
\rowcolor{PaperTableResult}
\ours & \TableBest{64.7} & \TableBest{13.0} & \TableBest{32.0} & 7.6 & \TableBest{22.6} & 2.9\\
\rowcolor{PaperTableStripe}
\ours-SC (E3) & \TableSecond{63.2} & 8.8 & 26.7 & 6.1 & 19.0 & 4.1\\
\bottomrule
\end{tabular}}
\end{minipage}
\hfill
\begin{minipage}[t]{\columnwidth}
\centering\small
\setlength{\tabcolsep}{2.5pt}
\caption{\textbf{Transplant coverage by model.} Cond. measures donor following on the $n$ pairs with both original answers correct. Dashes mark $n<10$. \Cref{fig:regional} uses common pairs.}
\label{tab:regional_coverage}
\resizebox{\linewidth}{!}{\begin{tabular}{l r rr r r}
\toprule Method & Full $N$ & Don. & Ctx. & $n/N$ & Cond. \\ \midrule
\rowcolor{PaperTableStripe}
SlideChat$^\dagger$ & 1349 & 1.4 & 2.9 & 57/1349 & 5.3\\
\rowcolor{white}
SLICEChat$^\dagger$ & 1124 & 6.9 & 7.9 & 296/1124 & 12.5\\
\rowcolor{PaperTableStripe}
HistoSelect$^\dagger$ & 1349 & 2.2 & \TableSecond{1.1} & 20/1349 & 10.0\\
\rowcolor{white}
Mean pooling & 1349 & 4.5 & 8.5 & 191/1349 & 2.1\\
\rowcolor{PaperTableStripe}
TCP-LLaVA$^\ast$ & 1349 & 0.0 & \TableBest{0.1} & 2/1349 & --\\
\rowcolor{white}
LoC-Path$^\ast$ & 1349 & 7.0 & 14.6 & 345/1349 & 9.3\\
\rowcolor{PaperTableStripe}
LoC-Path + \ours & 1349 & \TableSecond{29.6} & 8.7 & 541/1349 & 39.9\\
\rowcolor{white}
PathSelect$^\ast$ & 1349 & 11.7 & 6.4 & 319/1349 & 32.6\\
\rowcolor{PaperTableStripe}
TC-SSA-32 & 1349 & 7.5 & 12.2 & 287/1349 & 10.1\\
\rowcolor{white}
TC-SSA-8 & 1349 & 8.6 & 10.5 & 355/1349 & 15.2\\
\rowcolor{PaperTableStripe}
\ours, slide relations & 1349 & 14.8 & 9.3 & 576/1349 & 23.6\\
\rowcolor{PaperTableResult}
\ours & 1349 & \TableBest{32.2} & 7.3 & 535/1349 & 58.5\\
\rowcolor{PaperTableStripe}
\ours-SC (E3) & 1349 & 26.4 & 6.4 & 555/1349 & 50.5\\
\bottomrule\end{tabular}}
\end{minipage}
\end{table*}

This section evaluates spatial tumour selectivity and external cohort transfer. \Cref{sec:app_regional_ctrl} details the matched regional feature-exchange controls and independent pathologist validation of the CONCH tissue locator, while \cref{sec:app_transfer} evaluates zero-shot and adapted transfer to BCNB core biopsies and WSI-VQA without invoking the locator at inference.

\subsection{Regional Feature-Exchange Controls}
\label{sec:app_regional_ctrl}
Regional Transplant holds recipient surrounding tissue and exchanged patch count fixed while swapping tumour-associated patch features between contrastive patients. \Cref{fig:regional}b evaluates the common pairs shared across all compared models, whereas \cref{tab:transplant_controls} evaluates control pairs supporting matched tumour, non-tumour, and same-label exchanges on identical slides. Across \cref{tab:evidence_detailed,tab:regional_coverage}, Keep measures post-swap accuracy under same-label replacement, Selectivity subtracts the same-label flip rate from the different-label flip rate, Don.\ and Ctx.\ require both exchange directions to follow the donor or recipient label, and Deletion measures flips after removing tumour or context patches.

\paragraph{Locator calibration, pathologist validation, and donor specificity.}
Each patch is scored by CONCH~\citep{Lu2024VisualLanguage} against nine five-template prompt ensembles (tumour histology, organ epithelium, stroma, necrosis, lymphocytes, adipose, muscle, blood, and mucin) calibrated over 400 reference slides. A patch is assigned to tumour when its calibrated log-probability margin over the runner-up is above 1.0 nat, requiring at least 64 tumour and 64 non-tumour patches per slide (excluding glioma, sarcoma, clear-cell renal carcinoma, and lymphoma; deterministic test masks transfer to SLICEChat's grid where available). Against independent TCGA pathologist frozen-section estimates, within-cancer Spearman correlations are positive across 25 of 32 cancer types for tumour fraction and 29 of 31 for necrosis. These correlations validate locator fidelity outside our training pipeline. On matched control pairs (\cref{tab:transplant_controls,tab:regional_coverage}), \ours-8 follows donor labels far more often under tumour exchange than under non-tumour exchange at identical patch counts, preserves high joint consistency under same-label tumour replacement, and leads conditional donor following on correct pairs.

\subsection{External Transfer to BCNB and WSI-VQA}
\label{sec:app_transfer}
\begin{table}[!htbp]\centering\small
\setlength{\tabcolsep}{6pt}
\caption{\textbf{Which WSI-VQA* answers become consistent?} Each question counts once; Pass requires all four cyclic orders correct.}
\label{tab:wsivqa_transitions}
\begin{tabularx}{\linewidth}{Lrr}\toprule
Before adaptation & \multicolumn{2}{c}{After adaptation} \\
\cmidrule(lr){2-3} & Fail & Pass \\ \midrule
\rowcolor{PaperTableStripe} Fail & 39 & 13 \\
\rowcolor{white} Pass & 1 & 22 \\
\bottomrule\end{tabularx}\end{table}%

Zero-shot transfer in \cref{tab:auxiliary} evaluates BCNB core biopsies~\citep{Chen2025SlideChat,Xu2021BCNB} (averaged over cyclic orders across two-, three-, and four-option items) and WSI-VQA*~\citep{Chen2024WSIVQA} (closed-form four-option items with CONCH features), excluding all test patients from every training pool. Starting from the \cref{tab:main} \ours-8 checkpoint, two-epoch adaptation on disjoint WSI-VQA training patients with replayed Evidence items (selecting epoch~2 on a held-out validation split) converts a large group of previously failing WSI-VQA* questions into all-orders passes while losing only one prior pass (\cref{tab:wsivqa_transitions}), producing clear gains in both mean and all-orders accuracy (\cref{tab:auxiliary}) without invoking the locator at inference.

\section{Patient Weighting and Slot Restoration}
\label{sec:app_weighting}
This section reports two sensitivity disaggregations that test whether aggregate gains hold across individual patients and diagnostic attributes. \Cref{sec:app_pair_weight} checks SlideBench-Evidence paired correctness under patient-level reweighting (\cref{tab:patient_weight}), while \cref{sec:app_slot_restore} disaggregates pooled three-slot restoration across all 13 target attributes (\cref{fig:app_attribute_restore}).

\subsection{Patient-Level Pair Reweighting}
\label{sec:app_pair_weight}
In \cref{tab:main}, each SlideBench-Evidence question selects at most one same-cancer, different-label partner slide while retaining both original prompts and slides. Because distinct attributes can query the same slide, uniform pair averaging can weight multi-attribute patients more heavily. \Cref{tab:patient_weight} compares uniform directed-pair weighting against macro-averaging equally across unique source patients or unique donor patients. All three aggregations agree closely, shifting paired correctness by under one percentage point, and within \ours-8, single-sided outcomes remain the largest non-paired bucket while only jointly correct pairs earn credit.

\begin{table}[!htbp]
\centering\small
\setlength{\tabcolsep}{6pt}
\caption{\textbf{PathSleuth-8 paired correctness under patient reweighting.} All three aggregations evaluate the same 3,811 directed contrastive question pairs on SlideBench-Evidence.}
\label{tab:patient_weight}
\begin{tabularx}{\linewidth}{Lrr}
\toprule
Equal weight assigned per & Distinct units & Both correct (\%)\\
\midrule
\rowcolor{PaperTableStripe}Directed question pair & 3,811 & 33.3\\
\rowcolor{white}Source patient & 1,393 & 32.5\\
\rowcolor{PaperTableStripe}Donor patient & 1,095 & 32.7\\
\bottomrule
\end{tabularx}
\end{table}

\subsection{Attribute-Wise Slot Restoration}
\label{sec:app_slot_restore}
\begin{figure}[!htbp]
\centering
\includegraphics[width=0.92\linewidth]{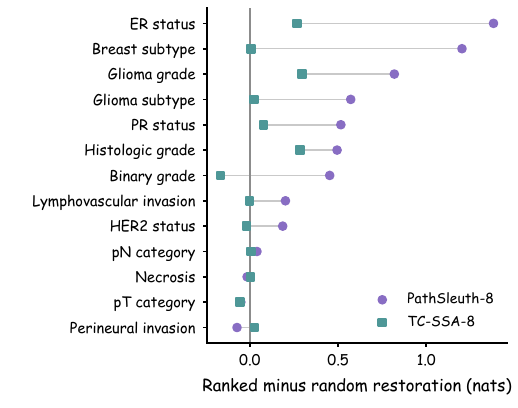}
\caption{\textbf{Per-attribute gain from probe-ranked three-slot restoration.} Mean paired log-probability recovery (probe-ranked minus random three slots, in nats) over 60 questions per attribute.}
\label{fig:app_attribute_restore}
\end{figure}

To disaggregate the pooled three-slot restoration curve in \cref{fig:vet_cumulative}, \cref{fig:app_attribute_restore} compares restoring the three highest inner-fold probe-AUC original slots into a full donor representation against restoring three randomly ordered slots across all 13 attributes (60 paired interventions per attribute sharing donor slides, option orders, and labels). Probe-ranked restoration outperforms random order on most attributes in \ours-8, with the clearest gains on ER and PR status, breast histologic subtype, and glioma grade and subtype, while matching random order within sampling noise on necrosis, pT category, and perineural invasion. Across receptor, grading, and subtype tasks, these recoveries show that EvA aligns reader sensitivity with compressed slots carrying H\&E-assessable correlates of each report label, complementing the patch-deletion tests in \cref{sec:patch_attribution}.

\end{document}